\documentclass[journal]{IEEEtran}
\usepackage{cite}
\usepackage{amsmath,amssymb,amsfonts}
\usepackage{algorithmic}
\usepackage{graphicx}
\usepackage{textcomp}
\usepackage{multicol}
\usepackage{balance}
\usepackage{soul}
\usepackage{multirow}
\usepackage{makecell}
\usepackage[table,xcdraw]{xcolor}
\usepackage{tikz}
\usetikzlibrary{shapes}
\usepackage[hyphens]{url}
\usepackage{multirow}
\def\BibTeX{{\rm B\kern-.05em{\sc i\kern-.025em b}\kern-.08em
    T\kern-.1667em\lower.7ex\hbox{E}\kern-.125emX}}

\begin{document}

\title{GPT (Generative Pre-trained Transformer) – A Comprehensive Review on Enabling Technologies, Potential Applications, Emerging Challenges, and Future Directions}

\author{Gokul~Yenduri, Ramalingam~M, Chemmalar~Selvi~G, Supriya~Y, Gautam~Srivastava, Praveen~Kumar~Reddy~Maddikunta, Deepti~Raj~G, Rutvij~H~Jhaveri, Prabadevi~B, Weizheng~Wang, Athanasios~V. Vasilakos, and Thippa~Reddy~Gadekallu

\thanks{Gokul~Yenduri, Ramalingam M, Chemmalar Selvi G, Supriya Y, Praveen Kumar Reddy Maddikunta, Deepti Raj G, Prabadevi B are with the School of Information Technology and Engineering, Vellore Institute of Technology, Vellore, Tamil Nadu- 632014, India (Emails: \{ gokul.yenduri, ramalingam.m, chemmalarselvi.g, supriya.d, praveenkumarreddy, deeptiraj.g2020, prabadevi.b \}@vit.ac.in)}

\thanks{Gautam Srivastava is with the Dept. of Math and Computer Science, Brandon University, Canada, and the Research Centre for Interneural Computing, China Medical University, Taichung, Taiwan as well as Dept. of Computer Science and Math, Lebanese American University, Beirut, Lebanon (email: srivastavag@brandonu.ca)}

\thanks{Rutvij H Jhaveri is with the Department of Computer Science and Engineering, School of Technology, Pandit Deendayal Energy University, India, (Email: rutvij.jhaveri@sot.pdpu.ac.in).}

\thanks{Weizheng Wang is with the Department of Computer Science, City University of Hong Kong, Hong Kong SAR, China, (E-mail: weizheng.wang@ieee.org).}

\thanks{Athanasios V. Vasilakos is with the Center for AI Research (CAIR),University of Agder(UiA), Grimstad, Norway, (Email: thanos.vasilakos@uia.no).}

\thanks{Thippa Reddy Gadekallu is with the School of Information Technology and Engineering, Vellore Institute of Technology, Vellore 632014, India, Lovely Professional University, Phagwara, India, Department of Electrical and Computer Engineering, Lebanese American University, Byblos, Lebanon, Jiaxing University , Jiaxing 314001, China, Zhongda Group, China, 314312 (E-mail: thippareddy@ieee.org).}}

\maketitle

\IEEEpubidadjcol

\begin{abstract}
The Generative Pre-trained Transformer (GPT) represent a notable breakthrough in the domain of natural language processing, which is propelling us toward the development of machines that can understand and communicate using language in a manner that closely resembles that of humans. GPT is based on the transformer architecture, a deep neural network designed for natural language processing tasks. Due to their impressive performance on natural language processing tasks and ability to effectively converse, GPT have gained significant popularity among researchers and industrial communities, making them one of the most widely used and effective models in natural language processing and related fields, which motivated to conduct this review. This review provides a detailed overview of the GPT, including its architecture, working process, training procedures, enabling technologies, and its impact on various applications. In this review, we also explored the potential challenges and limitations of a GPT. Furthermore, we discuss potential solutions and future directions. Overall, this paper aims to provide a comprehensive understanding of GPT, enabling technologies, their impact on various applications, emerging challenges, and potential solutions.

\end{abstract}



\IEEEpeerreviewmaketitle

\begin{table}[t]
    \renewcommand{\arraystretch}{1.3}
    \caption{List of key acronyms only if it is repeated }
    \centering
    \begin{tabular}{|p{1.25cm}|p{5.65cm}|}
        \hline
        \textbf{Acronyms} & \textbf{Description}\\
        
        \hline
        {AI} & Artificial Intelligence \\ 
        \hline
        {AR} & Augmented Reality 
        \\ 
        \hline
        {BERT} & Bidirectional Encoder Representations from Transformers 
        \\
        \hline
        {BGN} & Boneh–Goh–Nissim
        \\ 
        \hline
        {CNN} & ConvolutionalNeural Network 
        \\ 
        \hline
        {DAP} & Data Access Point \\ 
        \hline
        {DLT} & Decentralized Ledger Technology \\ 
        \hline
        {DL} & Deep Learning 
        \\ 
        \hline
        {DRL} & Deep Reinforcement Learning 
        \\ 
        \hline
        {DR} & Demand response 
        \\ 
        \hline
        {EC} & Edge Computing 
        \\ 
        \hline
        {EU} & End User 
        \\ 
        \hline
        {EAPs} & Energy Access Points 
        \\ 
        \hline
        {5G} & Fifth-Generation \\ 
        
        \hline
        {4G} & Fourth-Generation \\ 
        \hline
        {GPT} & Generative Pre-trained Transformer \\ 
        \hline
        {GPU} & Graphics Processing Unit \\ \hline
        {HPC} & High Performance Computing 
        \\ 
        \hline
        {HCI} & Human Computer Interaction \\ 
        \hline
        {IoT} & Internet of Things 
        \\ 
        \hline
        {ML} & Machine Learning 
        \\ 
        \hline
        {NLP} & Natural Language Processing \\ 
        \hline
        {NPC} & Non Playable Character 
        \\ 
        \hline
        {PLM} & Pre-trained Language Models 
        \\ 
        \hline
        {PTM} & Pre-Trained Models 
        \\ 
        \hline
        {RNN} & Recurrent Neural Network 
        \\ 
        \hline
        {6G} & Sixth-Generation \\ 
        \hline
        {TL} & Transfer Learning 
        \\ 
        \hline
        {VU} & Virtual Reality 
        \\ 
        \hline
              
    \end{tabular}
    \label{Tab:acronym}
\end{table} 
\section{Introduction} 
Language is the cornerstone of human communication and plays a vital role in shaping our interactions with the world. With the advent of NLP, it has revolutionized the way we interact with machines. NLP has become a game-changer in the world of communication, enabling humans to interact with machines in a more natural way. The evolution of NLP has been fueled by the exponential growth of textual data in the internet. Over the years, NLP has witnessed a significant transformation from simple rule-based systems to complex deep learning-based models. Despite the advances, natural language understanding and generation have long been a challenging problem in the field of NLP, largely due to the complex nature of human language. However, recent advancements have paved the way for the new approaches to tackle these challenges. One such breakthrough in NLP, is the development of the GPT \cite{han2021pre}. GPT became famous after the launch of ChatGPT by OpenAI, a research company \cite{OAI} that focuses on developing AI technologies. GPT is a deep learning model that is pre-trained on large corpora of text data and can be fine-tuned for specific tasks like language generation, sentiment analysis, language modelling, machine translation, and text classification. The transformer architecture used in GPT is a significant advancement over previous approaches to NLP, such as RNN and CNN. It uses a self-attention mechanism to allow the model to consider the context of the entire sentence when generating the next word, which improved the model's ability to understand and generate language. The decoder is responsible for generating the output text based on the input representation \cite{dong2018speech}.

GPT can perform a wide range of tasks in NLP. One of its key strengths is in natural language understanding (NLU), where it can analyze and comprehend the meaning of text, including identifying entities and relationships in sentences. It's also proficient in natural language generation (NLG), which means it can create text output, such as writing creative content or answering questions in a comprehensive and informative way. Alternatively, GPT is also code generator, where it can write programming code in various languages, such as Python or JavaScript. GPT can also be utilized for question answering, which means it can provide summaries of factual topics or create stories based on the input text. Additionally, GPT can summarize a piece of text, such as providing a brief overview of a news article or research paper, and it can be used for translation, which makes it possible to translate text from one language to another. Overall, GPT's ability to perform a wide range of NLP tasks with high accuracy and precision, makes it an invaluable tool for various industries, including finance, healthcare, marketing, and more. As NLP technology continues to advance, we can expect GPT and other language models to become even more sophisticated and powerful, enabling us to communicate with machines more naturally and effectively.

\subsection{Motivation}

GPT has become a transformative technology in the field of NLP, enabling the rapid development and growth of a wide range of industries and applications. Despite its wide adoption and numerous potential applications, there is still much to be explored and understood about GPT's capabilities. Although there are studies on GPT in the literature related to academia and libraries \cite{lund2023chatting}, education \cite{kasneci2023chatgpt},
GPT models\cite{qiu2020pre}, banking and corporate communication \cite{george2023revolutionizing}, advancements in chatGPT and its version \cite{zhang2023complete}, and on generative AI's \cite{zaib2020short}, no existing reviews are dedicated to providing a comprehensive survey on GPT. Therefore, there is a need for a comprehensive review that focuses on GPT's architecture, enabling technologies, potential applications, emerging challenges, interesting projects and future directions. These limitations motivated us to conduct this review. Hence, this review will not only help researchers and practitioners in this field to gain a better understanding of GPT but also provide valuable insights into its potential applications and major limitations when conducting the research.

\subsection{Related Surveys and Contributions}

The GPT model is a type of DL model that uses self-supervised learning to pre-train massive amounts of text data, enabling it to generate high-quality language output. The recent advancements in GPT model research can be attributed to the continual improvement of its architecture, increased availability of computing power, and the development of novel techniques to fine-tune the model for specific tasks. These advancements have led to the creation of larger and more powerful GPT models, enabling them to perform a wider range of NLP tasks with unprecedented accuracy and fluency. These GPT models have demonstrated great potential in transforming various industries like healthcare \cite{liu2023deid}, customer service \cite{rivas2023marketing}, financial industry \cite{leippold2023thus} and so on. These applications are enabled by the generation of high-quality and diverse data like large-scale corpora of text data with different fast-growing enabling technologies\cite{trajtenberg2018ai,haluza2023artificial}. There are numerous survey papers published to provide a comprehensive overview of the latest developments in GPT models, insights into the different architectures, training methods, evaluation metrics, and highlight the challenges and future directions of this field. This literature survey aims to review and analyze the key findings and contributions of the most recent survey papers published on GPT models, to provide a comprehensive and up-to-date understanding of the state-of-the-art in this exciting and rapidly evolving field.


Lund et al. \cite{lund2023chatting} presents the potential effects of AI and GPT models, specifically ChatGPT, on academia and libraries. They discussed the capabilities of ChatGPT in generating human-like responses and its potential applications. They examine how AI-powered chatbots and virtual assistants based on GPT models can enhance student learning experiences, assist with research tasks, and improve library services. They also address concerns regarding data privacy, biases, and the need for ethical guidelines. Overall, this survey paper highlighted the transformative potential of AI and GPT models while emphasizing the importance of responsible deployment and human oversight.

Kasneci et al. \cite{kasneci2023chatgpt} have reviewed the potential opportunities and challenges of using large language models, specifically ChatGPT, for educational purposes. They highlighted the benefits and limitations of using such models by discussing their implications for teaching and learning. In addition, a defined strategy and pedagogical approach with a heavy focus on critical thinking and fact-checking are required while using such large language models in educational institution. Thus, they concluded the paper by highlighting the key technical challenges like copyright issues, biased content creation, user dependency, privacy and security, and high-cost language models when such language models are used in the educational sector.

Qiu et al. \cite{qiu2020pre} presented an exhaustive survey of various types of GPT models by detailing their working architecture. They discussed the evolution of pre-training methods for NLP, from language modelling to TL and pre-training on large-scale corpora. It also reviews the different types of GPT models, including word embeddings, contextual embeddings, and transformer-based models, and discusses their applications in various NLP tasks such as text classification, Named Entity Recognition, and machine translation. They highlighted the benefits of GPT's models for the NLP domain, such as its ability to improve model performance with limited annotated data, reduce the need for task-specific feature engineering, and enable TL across multiple tasks. They discussed the major challenges and limitations of PTMs, such as the risk of bias and the lack of interpretability.

George et al. \cite{george2023revolutionizing} studied the potential impact of GPT-4, the next iteration of GPT models, on communication within corporate environments. They discussed how GPT-4 can revolutionize business communication by enabling more efficient and effective interactions. They explore various applications of GPT-4 in corporate settings, such as automating customer support through AI chatbots that can provide personalized responses and resolve queries in real-time. They also addressed potential challenges and considerations associated with implementing GPT-4 in corporate settings. These include concerns about data security, privacy, and the need for human oversight to ensure accurate and ethical communication. Thus, they concluded by emphasizing the transformative potential of GPT-4 in revolutionizing business communication to fully harness the benefits of GPT-4 while addressing any potential risks or limitations.


Zhang et al. \cite{zhang2023complete} presents an extensive survey of generative AI and evaluates the capabilities of the ChatGPT models, particularly from GPT-4 to GPT-5. They provided an overview of generative AI, highlighting its significance in generating realistic and creative outputs across various domains and evaluate their advancements over previous iterations. They analyze the architectural improvements, model size, training techniques, and dataset considerations employed in GPT-4 and GPT-5. In addition to it, they presented a comprehensive comparison of ChatGPT with other state-of-the-art generative AI models, such as OpenAI's DALL-E and CLIP. Finally, they concluded with valuable insights into the capabilities and limitations of these models and highlights the broader landscape of generative AI.


Zaib et al. \cite{zaib2020short} provides a survey on the latest advancements in GPTS and PTMs for conversational AI applications. They focused on PLMs and their approaches while building dialogue-based systems. They also highlighted the potential use of transformer-based models such as BERT and GPT, which have demonstrated good performance in understanding NLP generation, and dialogue management. Thus, they concluded with the significant challenges in the field of developing conversational AI systems using PLMs and GPTs.

\begin{figure*}[!ht]
	\centering
	\includegraphics[width=0.8\textwidth]{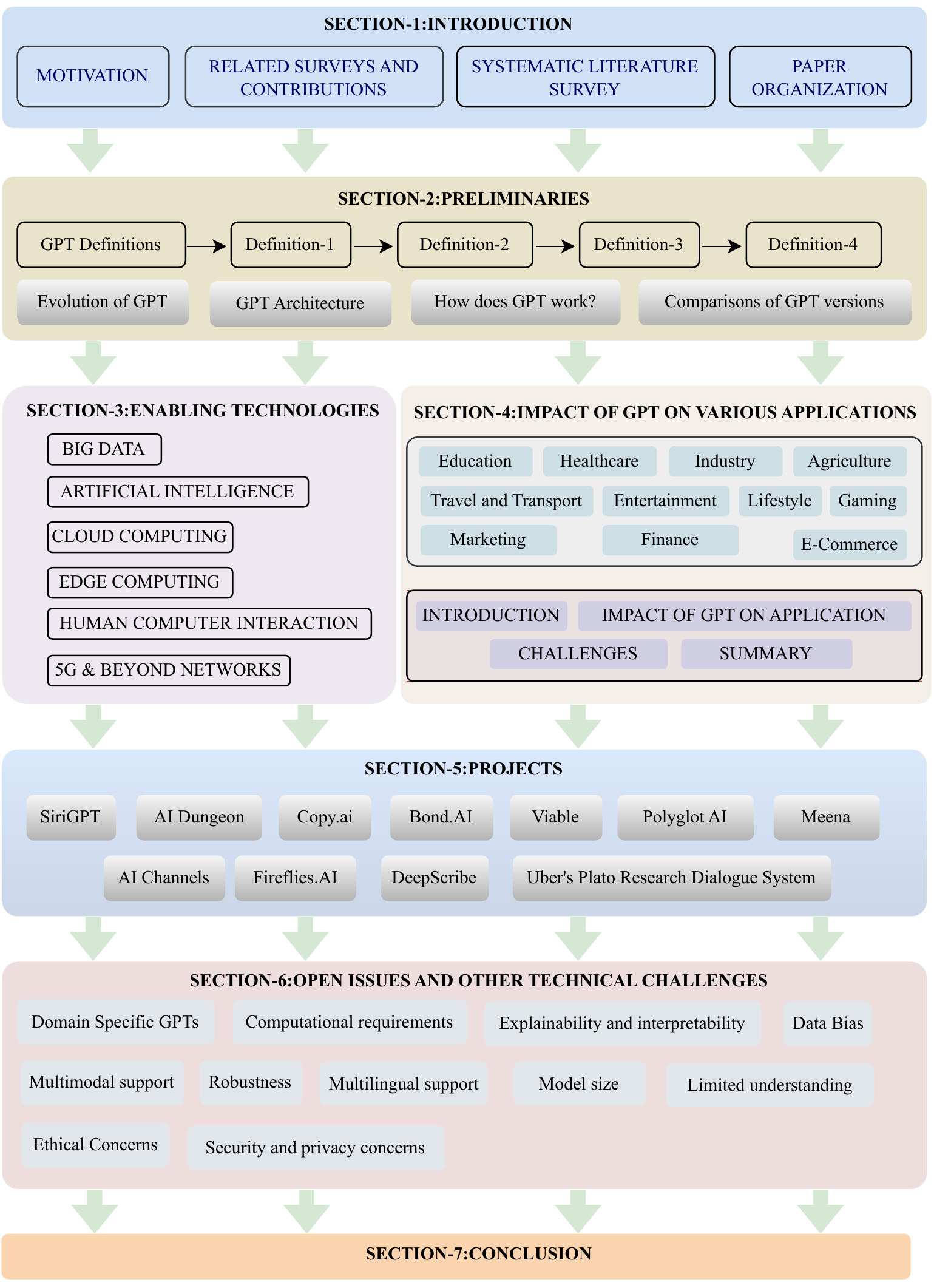}
	\caption{Organization Chart of the survey.}
	\label{fig:Organization chart}
\end{figure*}

Thus, the comparison of existing surveys on GPT models highlighting the growing importance of these models in key areas of NLP and other related fields are discussed here. Hence, this is the first-of-its-kind survey that presents the extensive information, by comparing existing surveys with our survey and summarized in Table \ref{tab:RelatedWorkComparison}.


\begin{table*}[!ht]
\centering
\caption{Comparison of this survey with the existing surveys}
\label{tab:RelatedWorkComparison}
\resizebox{\textwidth}{!}{%
\begin{tabular}{|p{1.5cm}|p{1 cm}|p{1 cm}|p{1 cm}|p{1 cm}|p{1 cm}|p{1 cm}|p{1 cm}|p{1 cm}|p{1 cm}|p{1 cm}|p{1 cm}||p{1 cm}|p{1 cm}|p{1 cm}|p{1 cm}|p{1 cm}|p{1 cm}||p{6 cm}|}
\hline
\multirow{2}{*}{\textbf{Ref.}} &
  \multicolumn{11}{l|}{\textbf{Applications}} &
  \multicolumn{6}{l|}{\textbf{Enabling Technologies}} &
  \multirow{2}{*}{\textbf{Remarks}} \\ \cline{2-18}
 &
  \multicolumn{1}{l|}{\rotatebox[origin=c]{90}{~Education~}} &
  \multicolumn{1}{l|}{\rotatebox[origin=c]{90}{~Industry~}} &
  \multicolumn{1}{l|}{\rotatebox[origin=c]{90}{~Agriculture~}} &
  \multicolumn{1}{l|}{\rotatebox[origin=c]{90}{~Healthcare~}} &
  \multicolumn{1}{l|}{\rotatebox[origin=c]{90}{~Transport~}} &
  \multicolumn{1}{l|}{\rotatebox[origin=c]{90}{~E-Commerce ~}} &
  \multicolumn{1}{l|}{\rotatebox[origin=c]{90}{~Entertainment~}} &
  \multicolumn{1}{l|}{\rotatebox[origin=c]{90}{~Lifestyle~}} &
  \multicolumn{1}{l|}{\rotatebox[origin=c]{90}{~Gaming~}} &
  \multicolumn{1}{l|}{\rotatebox[origin=c]{90}{~Marketing~}} &
  \rotatebox[origin=c]{90}{~Finance~} &
  \multicolumn{1}{l|}{\rotatebox[origin=c]{90}{~Big Data ~}} &
  \multicolumn{1}{l|}{\rotatebox[origin=c]{90}{~AI~}} &
  \multicolumn{1}{l|}{\rotatebox[origin=c]{90}{~Cloud Computing ~}} &
  \multicolumn{1}{l|}{\rotatebox[origin=c]{90}{~Edge Computing~}} &
  \multicolumn{1}{l|}{\rotatebox[origin=c]{90}{~5G and Beyond~}} &
  {\rotatebox[origin=c]{90}{~HCI~}} &
   \\ \hline
\cite{lund2023chatting} &
  \multicolumn{1}{l|}{$\checkmark$} &
  \multicolumn{1}{l|}{X} &
  \multicolumn{1}{l|}{X} &
  \multicolumn{1}{l|}{X} &
  \multicolumn{1}{l|}{X} &
  \multicolumn{1}{l|}{X} &
  \multicolumn{1}{l|}{X} &
  \multicolumn{1}{l|}{X} &
  \multicolumn{1}{l|}{X} &
  \multicolumn{1}{l|}{X} &
  X &
  \multicolumn{1}{l|}{X} &
  \multicolumn{1}{l|}{$\checkmark$} &
  \multicolumn{1}{l|}{$\checkmark$} &
  \multicolumn{1}{l|}{$\checkmark$} &
  \multicolumn{1}{l|}{X} &
  X & They conducted a survey discussing capabilities of ChatGPT on academia and
libraries. Although, Key challenges of Chatgpt were highlighted, practical implementation challenges and research directions were missing.
   \\ 
\hline
      
\cite{kasneci2023chatgpt} &
  \multicolumn{1}{l|}{$\checkmark$} &
  \multicolumn{1}{l|}{X} &
  \multicolumn{1}{l|}{X} &
  \multicolumn{1}{l|}{X} &
  \multicolumn{1}{l|}{X} &
  \multicolumn{1}{l|}{X} &
  \multicolumn{1}{l|}{X} &
  \multicolumn{1}{l|}{X} &
  \multicolumn{1}{l|}{X} &
  \multicolumn{1}{l|}{X} &
  X &
  \multicolumn{1}{l|}{X} &
  \multicolumn{1}{l|}{$\checkmark$} &
  \multicolumn{1}{l|}{X} &
  \multicolumn{1}{l|}{X} &
  \multicolumn{1}{l|}{X} &
  X &
  Reviewed the potential opportunities
and challenges of using large language models, specifically
ChatGPT, for educational purposes. Thus, evolution of GPT and their preliminaries were not discussed in this survey paper. \\ 
   \hline
   
\cite{qiu2020pre} &
  \multicolumn{1}{l|}{X} &
  \multicolumn{1}{l|}{X} &
  \multicolumn{1}{l|}{X} &
  \multicolumn{1}{l|}{X} &
  \multicolumn{1}{l|}{X} &
  \multicolumn{1}{l|}{X} &
  \multicolumn{1}{l|}{X} &
  \multicolumn{1}{l|}{X} &
  \multicolumn{1}{l|}{X} &
  \multicolumn{1}{l|}{X} &
  X &
  \multicolumn{1}{l|}{X} &
  \multicolumn{1}{l|}{$\checkmark$} &
  \multicolumn{1}{l|}{X} &
  \multicolumn{1}{l|}{X} &
  \multicolumn{1}{l|}{X} &
  X &
  They presented an exhaustive survey of various
types of GPT models by detailing their working architecture with benefits and limitations of GPTs. However,  \\ \hline

\cite{george2023revolutionizing} &
  \multicolumn{1}{l|}{X} &
  \multicolumn{1}{l|}{X} &
  \multicolumn{1}{l|}{X} &
  \multicolumn{1}{l|}{$\checkmark$} &
  \multicolumn{1}{l|}{X} &
  \multicolumn{1}{l|}{X} &
  \multicolumn{1}{l|}{X} &
  \multicolumn{1}{l|}{X} &
  \multicolumn{1}{l|}{X} &
  \multicolumn{1}{l|}{X} &
  X &
  \multicolumn{1}{l|}{X} &
  \multicolumn{1}{l|}{$\checkmark$} &
  \multicolumn{1}{l|}{X} &
  \multicolumn{1}{l|}{X} &
  \multicolumn{1}{l|}{X} &
  X &
  Studied the potential impact of GPT4 in business communication and explore various applications of GPT-4 in corporate settings by highlighting any potential
risks or limitations. But, how GPT architecture can be used in corporate is not found with key enabling technologies.
 \\ \hline
\cite{zhang2023complete} &
  \multicolumn{1}{l|}{X} &
  \multicolumn{1}{l|}{$\checkmark$} &
  \multicolumn{1}{l|}{X} &
  \multicolumn{1}{l|}{X} &
  \multicolumn{1}{l|}{$\checkmark$} &
  \multicolumn{1}{l|}{$\checkmark$} &
  \multicolumn{1}{l|}{$\checkmark$} &
  \multicolumn{1}{l|}{X} &
  \multicolumn{1}{l|}{X} &
  \multicolumn{1}{l|}{$\checkmark$} &
  {$\checkmark$} &
  \multicolumn{1}{l|}{X} &
  \multicolumn{1}{l|}{$\checkmark$} &
  \multicolumn{1}{l|}{X} &
  \multicolumn{1}{l|}{X} &
  \multicolumn{1}{l|}{X} &
  X &
  Analyzed the architectural
improvements, model size, training techniques, and dataset
considerations employed in GPT-4 and GPT-5. However, preliminary details are unedr explored. \\ \hline
\cite{zaib2020short} &
  \multicolumn{1}{l|}{X} &
  \multicolumn{1}{l|}{X} &
  \multicolumn{1}{l|}{X} &
  \multicolumn{1}{l|}{X} &
  \multicolumn{1}{l|}{X} &
  \multicolumn{1}{l|}{X} &
  \multicolumn{1}{l|}{X} &
  \multicolumn{1}{l|}{X} &
  \multicolumn{1}{l|}{X} &
  \multicolumn{1}{l|}{X} &
  X &
  \multicolumn{1}{l|}{X} &
  \multicolumn{1}{l|}{X} &
  \multicolumn{1}{l|}{X} &
  \multicolumn{1}{l|}{X} &
  \multicolumn{1}{l|}{X} &
  X &
  Recent trends in   language models, applications of dialogue management, question answering NLP   tasks were discussed along with challenges and future scope of GPT. Although it covered most of the technical aspects, the integration challenges to overcomeare not presented. \\ \hline
Our Survey   Paper &
  \multicolumn{1}{l|}{$\checkmark$} &
  \multicolumn{1}{l|}{$\checkmark$} &
  \multicolumn{1}{l|}{$\checkmark$} &
  \multicolumn{1}{l|}{$\checkmark$} &
  \multicolumn{1}{l|}{$\checkmark$} &
  \multicolumn{1}{l|}{$\checkmark$} &
  \multicolumn{1}{l|}{$\checkmark$} &
  \multicolumn{1}{l|}{$\checkmark$} &
  \multicolumn{1}{l|}{$\checkmark$} &
  \multicolumn{1}{l|}{$\checkmark$} &
  $\checkmark$ &
  \multicolumn{1}{l|}{$\checkmark$} &
  \multicolumn{1}{l|}{$\checkmark$} &
  \multicolumn{1}{l|}{$\checkmark$} &
  \multicolumn{1}{l|}{$\checkmark$} &
  \multicolumn{1}{l|}{$\checkmark$} &
  $\checkmark$ &
  Presents the evolution of GPT models, GPT architecture and its detailed working, key enabling technologies, significant advancements of GPT models and their potential benefits in real-life applications, GPT projects, lessons learnt, open challenges and future research directions.
   \\ \hline
\end{tabular}%

}
\end{table*}

\subsection{Systematic Literature Survey}

In this review of GPT, we conducted a thorough literature review using various reputable sources. Our search was primarily focused on peer-reviewed journals, and high-quality articles from reputed national and international conferences, seminars, books, symposiums, and journals. To ensure the credibility of our sources, we referred to well-known archives such as Google Scholar and arXiv, and publications from top databases like IEEE, Springer, Elsevier, Taylor \& Francis, and Wiley. To identify relevant GPT references and publications, we used keywords such as NLP\-GPT, GPT architecture, DL for GPT, Pre\-training GPT, Fine-tuning AI GPT and GPT vertical applications. We then screened all the retrieved articles based on their titles, excluding any papers with poor-quality material. Next, we reviewed the abstracts of the remaining articles to determine their contributions. In the final step of our literature review, we extracted the necessary data for our analysis. By following these phases, we ensured that our study was based on high-quality and credible sources.

\subsection{Paper Organization}

The structure of this paper's organization is illustrated in Fig. \ref{fig:Organization chart}. Section 2 presents the preliminaries of GPT models such as the definition of GPT, its evolution and architecture, how it works and presents the comparison of various GPT models. Section 3 discusses the key enabling technologies for GPT models. The impact of GPT models in various applications are presented in Section 4. In Section 5, we highlighted some of the exciting GPT projects that are currently developed. Section 6 includes open issues, other technical challenges and future research directions in the field of GPT. Finally, we conclude the paper in Section 7, by summarizing the key findings and contributions of this study. The list of key acronyms are listed in Table \ref{Tab:acronym}. 

\section{Preliminaries}
In this section, the evolution of GPT models, the architecture of GPT, working process of GPT models are discussed and finally, different versions of GPT models are compared.
\subsection {Generative Pre-trained Transformer}
The GPT model produces enormous quantities of pertinent and complicated machine-generated text from a small amount of text as input. GPT models can be identified as a language model that mimics human text using a DL techniques and it acts as an autoregressive model in which the present value is based on the previous value \cite{quintans2023chatgpt}.
\subsubsection{Definition 1}
GPTs are language models pre-trained on vast quantities of textual data and can perform a wide range of language-related tasks \cite{zhu2022generative}.
\subsubsection{Definition 2}
A GPT is a language model relying on DL that can generate human-like texts based on a given text-based input.  \cite{work3}. 
\subsubsection{Definition 3}
GPT is a language model developed by OpenAI to help give systems intelligence and is used in such projects as ChatGPT \cite{work3}.
\subsection {Evolution of GPT}
GPT models have evolved through multiple changes and breakthroughs in NLP technology. These are some significant turning points in the growth of the GPT model:
\begin{figure*}[!ht]
	\centering
	\includegraphics[width=\linewidth]{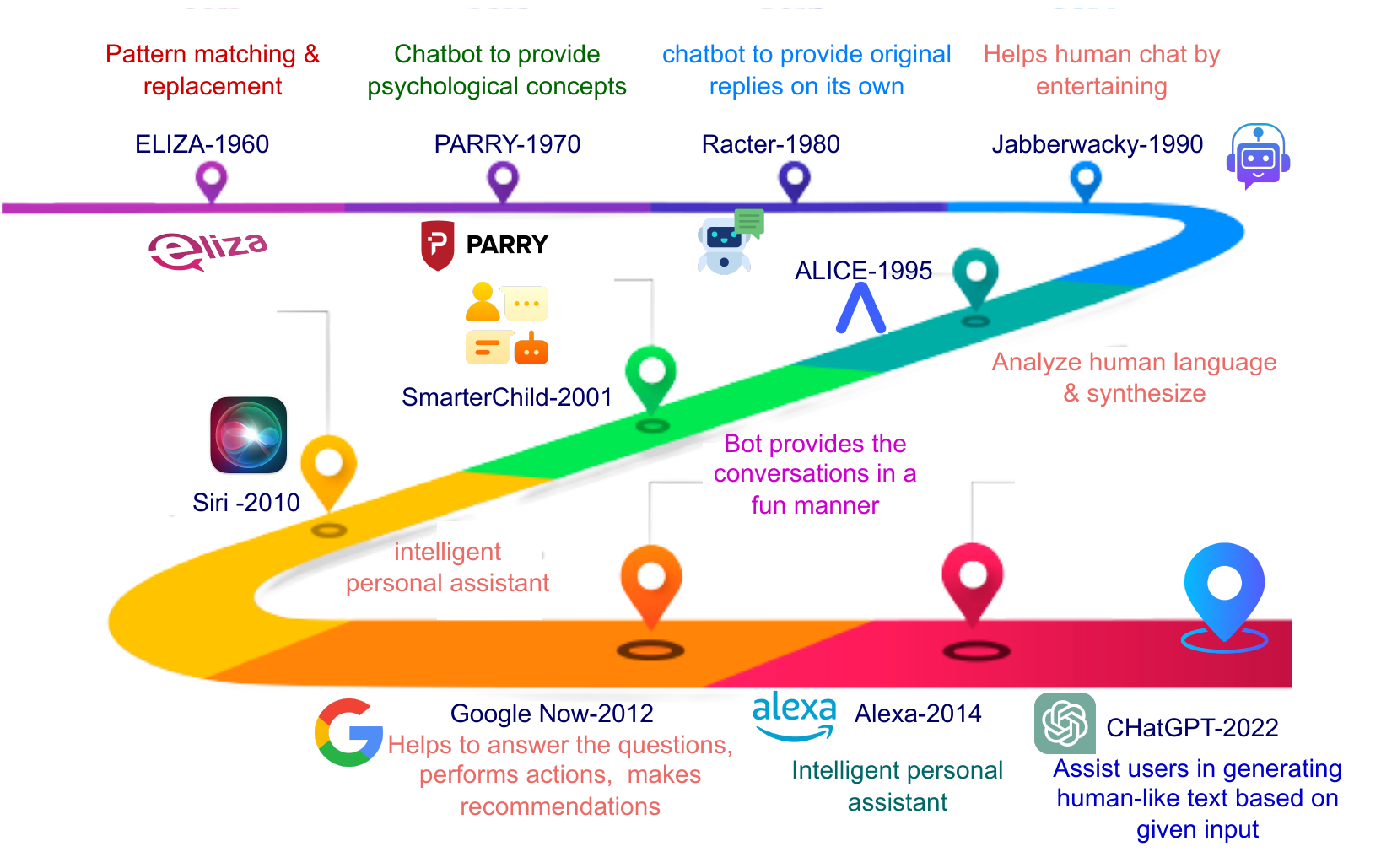}
	\caption{GPT Road Map.}
	\label{fig:origin}
\end{figure*}
Before GPT, NLP models have been trained on large amounts of annotated data that is related to a specific task. This had a significant drawback because it was difficult to access the quantity of labelled data required to train the model precisely. The NLP models were unable to complete tasks outside of their training set since they were restricted to a particluar set of data. To get around these restrictions, OpenAI offered a Generative Language Model called GPT-1 that was created using unlabeled data and then given to users to fine-tune to carry out subsequent tasks like sentiment analysis, categorization, and question-answering \cite{kosinski2023theory}. This indicates that the model attempts to produce an appropriate response based on input and that the data used to train the model is not labelled \cite{pre-work}. Fig. \ref{fig:origin} shows the timeline of the evolution of several pre-trained models from Eliza, which was created in 1960, to the more current 2022-ChatGPT.\\ 
GPT-1 was the first ever model that could read the text and respond to queries \cite{ghojogh2020attention}. OpenAI released GPT-1 in 2018. GPT-1 was a major move forward in AI development because it enabled computers to comprehend textual material in a more natural manner than before. This generative language model was able to learn a wide variety of connections and gain immense knowledge on a varied corpus of contiguous text and lengthy stretches \cite{work1}. This happened after being trained on a huge BooksCorpus dataset. In terms of design, GPT-1 employs a 12-layer decoder architecture transformer with a self-attention system for training. GPT-1's capacity to execute zero-shot performance on different tasks was one of its major success as a result of its pre-training. This ability demonstrated that generative language modelling can be used to generalize the model when combined with a successful pretraining idea. With TL as its foundation, GPT models evolved into a potent tool for performing NLP tasks with minimal fine-tuning \cite{williams2023algorithmic}. It paved the way for other models to progress even more in generative pre-training using larger datasets and parameters.  \cite{kosinski2023theory}.\\
To create a better language model later in 2019, OpenAI created a GPT-2 using a bigger dataset and more parameters. The model design and execution of GPT-2 are some of the key advancements \cite{9719940}. With 1.5 billion parameters, it has 10 times the size of GPT-1 (117 million parameters), and it has 10 times as many parameters and data \cite{work1}. By using only the raw text as input and utilizing little to no training examples, it is effective in terms of resolving various language tasks related to translation, summarization, etc. Evaluation of GPT-2 on various downstream task datasets revealed that it excelled by substantially increasing accuracy in recognizing long-distance relationships and predicting sentences \cite{taecharungroj2023can}. \\
The most recent iteration of the GPT model is GPT-3. It is a sizable language prediction and production model created by OpenAI that can produce lengthy passages of the source text. GPT-3 eventually emerged as OpenAI's ground-breaking AI language software. Simply put, it is a piece of software that can create lines on its own that are so distinctive they almost sound like they were written by a human \cite{spitale2023ai}. The GPT-3 program is presently accessible with limited access via a cloud-based API, and access is required to investigate the utility. Since its debut, it has produced several interesting apps. Its capacity, which is about 175 billion parameters big and 100 times larger than GPT-2, is a key advantage. It is taught using a corpus of 500 billion words called "Common Crawl" that was gathered from a sizable content archive and the internet \cite{ye2023comprehensive}. Its other noteworthy and unexpected capability is its ability to carry out basic mathematical operations, write bits of code, and carry out clever tasks. As a result, NLP models can help businesses by responding more quickly to requests and accurately keeping best practices while minimizing human mistakes \cite{liu2021makes}. Due to its intricacy and size, many academics and writers have referred to it as the ultimate black-box AI method. Due to the high cost and inconvenience of performing inference, as well as the billion-parameter size that makes it resource-intensive, it is difficult to put into practice in jobs \cite{taecharungroj2023can}. \\
GPT-4 was named as the successor of GPT-3. In the meantime, several AI models built on GPT-3.5, an updated version of GPT-3, have been surreptitiously released by OpenAI \cite{bommarito2022gpt}. GPT-3.5 was trained on a mixture of text and code. From the vast amounts of data collected from the web, which includes tens and thousand of Wikipedia entries, social media posts, and news items, GPT 3.5 learned the relations between words, sentences, and various components. It was utilized by OpenAI to develop several systems that have been tailored to complete particular jobs \cite{ye2023comprehensive}. It collected vast amounts of data from the web, including tens of thousands of Wikipedia entries, posts on social media, and news items, and used that information to learn the relationships between sentences, words, and word components \cite{hagendorff2022machine}.\\
The latest version of the GPT model by OpenAI is GPT-4 which is a multimodal big language model. It was launched on March 14, 2023, and is now accessible to the general public through ChatGPT Plus in a constrained capacity. A waitlist is required to gain access to the business API \cite{liu2023deid}. Using both public data and "data licensed from third-party providers," GPT-4 was pre-trained to anticipate the next coin as a transformer. It was then adjusted with reinforcement learning based on input from humans and AI for human alignment and policy conformance. In comparison to GPT-3, which had context windows of only 4096 and 2049 tokens, respectively, the group created two variants of GPT-4 with context windows of 8192 and 32768 tokens.
\subsection{GPT model's architecture}

GPT models are based on neural networks that are used for NLP tasks, such as language modelling, text classification, and text generation.\\
The GPT model's architecture is based on the transformer model \cite{vaswani2017attention}. The Transformer model uses self-attention mechanisms to process input sequences of variable length, making it well-suited for NLP tasks. GPT simplifies the architecture by substituting encoder-decoder blocks with decoder blocks. GPT model takes the transformer model and pre-trains it on large amounts of text data using unsupervised learning techniques. The pre-training process involves predicting the next word in a sequence given the previous words, a task known as language modelling. This pre-training process enables the model to learn representations of natural language that can be fine-tuned for specific downstream tasks \cite{hou2023geneturing}. The following are the components of the GPT architecture.\\
\begin{itemize}

   \item{Input Embedding layer:} The embedding layer maps the input tokens (e.g., words or subwords) to continuous vector representations, which can be processed by the transformer blocks\cite{edunov-etal-2019-pre}.
   \item{Positional encoding:} Since the transformer blocks do not have any notion of order or position, positional encoding is added to the input embeddings to provide information about the relative position of tokens.
Masking: In some cases, masking may be necessary to mask certain input tokens (e.g., in language modelling tasks, the model should only use tokens that come before the target token).
Transformer blocks: GPT models are based on the transformer architecture. It is designed for NLP tasks and has been widely used in applications such as machine translation, text classification, and text generation. Transformers allow the model to focus on different areas of the input while processing \cite{rahali2023end}.
   \item{Linear and Softmax Functions:}
In the GPT architecture, the softmax function is commonly used for classification tasks. The softmax function is applied to the output of the final layer of the model. It generates a probability distribution over a set of output classes. The output of the final layer is specifically converted into a set of logits before being normalized with the softmax function. The normalized values obtained from the model can be interpreted as the likelihood or probability that a particular input belongs to each of the output classes.
The query, key, and value vectors for each token in the input sequence are frequently calculated using linear functions in the attention mechanism. The output of the multi-head attention layer is transformed using them in the feedforward layers as well. The output layer also employs linear functions to forecast the following token in the sequence \cite{stevens2021softermax}.
   \item{Pre-training:} Pre-training is a key component of the GPT architecture. In pre-training, the model is trained on a large amount of data in an unsupervised manner even before fine-tuning the model for specific tasks like classification and text generation.
   \item{Fine-tuning:}
Fine-tuning is the process of adapting a pre-trained neural network model, such as GPT, to a new task or dataset by further training the model on that task or dataset. Fine-tuning in GPT involves adjusting the parameters of the pre-trained model to optimize performance on a specific downstream task, such as text classification or text generation \cite{bangura2023automatic}.
   \item{Language modeling:}
Language modelling is a key task in the GPT architecture. In the case of GPT, the language modelling task is performed during the pre-training phase of the model. In pre-training, the model is trained based on a large amount of data using a language model objective. It is the task of predicting the next word in sequence based on the previous words. It allows the model to learn relationships between the words and their meaning in the training data \cite{savelka2023large}.
   \item{Unsupervised learning:}
Unsupervised learning is an ML algorithm which enables the model to learn form unlabelled data without any human intervention. GPT models use unsupervised learning in the pre-training phase to understand the relationships between the words and their context in the training data \cite{liu2023variational}. 

\end{itemize}

\begin{table*}[!ht]
\centering
\caption{Comparsion of different versions of GPT model}
\label{tab:comparisions}
\resizebox{\textwidth}{!}{%
\begin{tabular}{|l|p{2cm}|p{2cm}|p{2cm}|p{2cm}|l|p{2.7cm}|p{2.5cm}|p{3.5cm}|}
\hline
Model &
  Tokens &
  Size &
  Parameters &
  Dataset &
  Year &
  Features &
  Input Type &
  Drawbacks \\ \hline
GPT-1 &
  - &
  12-layer decoder &
  117M parameters &
  Books corpus &
  2018 &
  Used   mostly for language modelling tasks and it is transformer based &
  A sequence of tokens and words &
  Limited   Capacity, Limited Data, Cannot perform complex tasks, Limited applications \\ \hline
GPT-2 &
  - &
  10 times the size of   GPT-1 &
  1.5B parameters &
  Downstream task   datasets &
  2019 &
  Text   generation capabilities are improved and a chance for misuse &
  A sequence of tokens and words &
  Limited   Control, Limited Data Diversity, Expensive computational requirements, Risk of   improper information \\ \hline
GPT-3 &
  4096 and 2049 tokens &
  100 times larger than   GPT-2 &
  175B parameters &
  Common Crawl &
  2020 &
  Good   NLP capabilities, language translation, summarization and generation of text &
  A sequence of tokens and words and images and tables &
  Limited   Control, Limited Data Diversity, Lack of explanation, Ethical concerns \\ \hline
GPT-3.5 &
   maximum token limit of 4096 tokens &
  96 layers &
  similar or larger number of parameters like GPT-3 &
  - &
  2022 &
  Improves user experience by delivering more precise and contextually relevant information &
  The input type typically consists of text data &
  Limited resources to train,Data Bias,Lack of Explainability,Limited Contextual Understanding,High Inference Latency \\ \hline
GPT-4 &
  8192 and 32768 tokens &
  - &
  100T parameters &
  - &
  2023 &
  Creative   and technical writing tasks &
  A sequence of tokens and words and images and tables &
  - \\ \hline
\end{tabular}%
}
\end{table*}
\subsection{How do GPT models work?}

\begin{figure}[!ht]
	\centering
	\includegraphics[width=\linewidth]{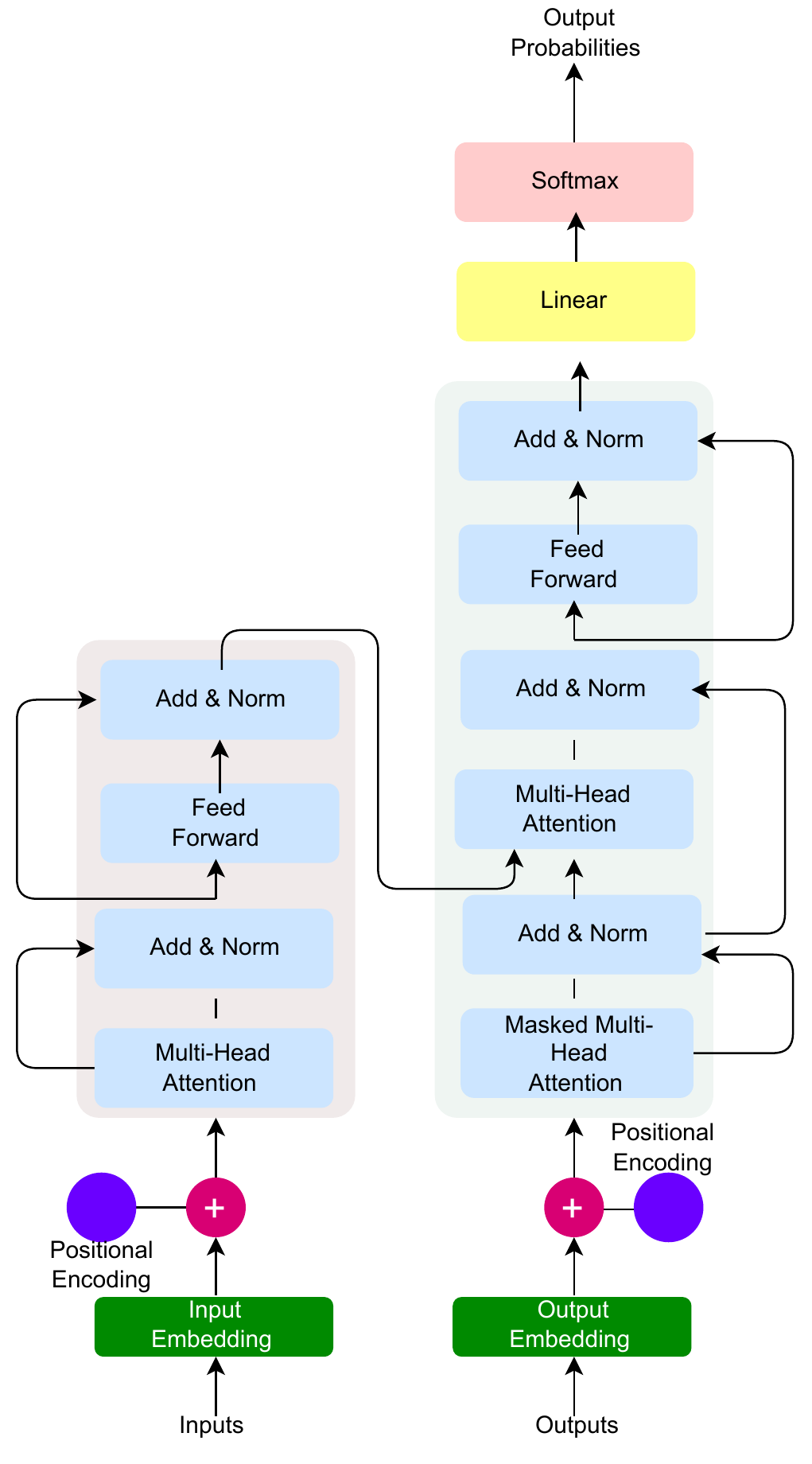}
	\caption{Transformer Architecture.}
	\label{fig:transformer}
\end{figure}

\begin{figure*}[!ht]
	\centering
	\includegraphics[width=\linewidth]{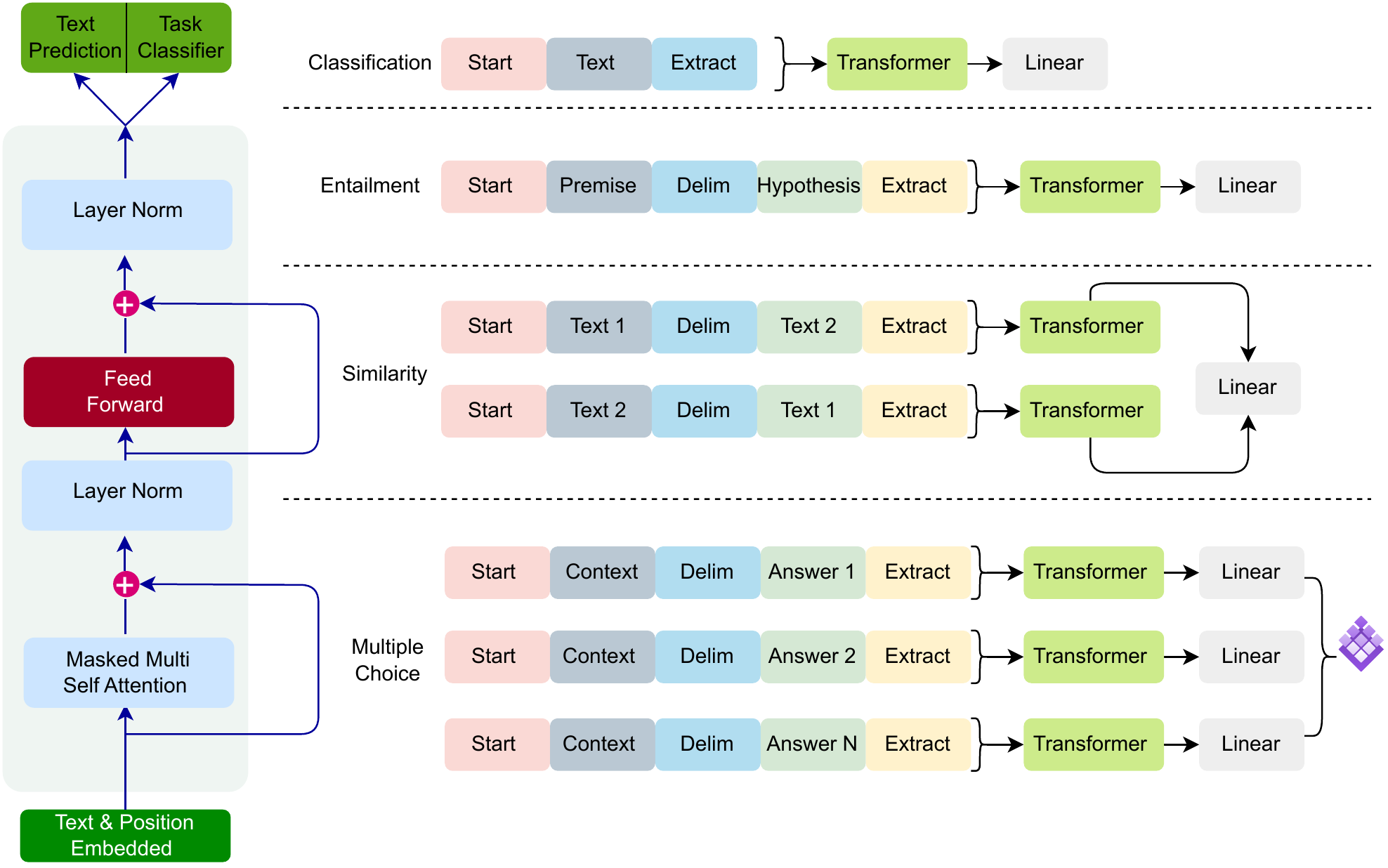}
	\caption{Transformer Architecture and Input Transformations for Fine-Tuning on Different Tasks.}
	\label{fig:Inputtransformer}
\end{figure*}

\begin{figure*}[!ht]
	\centering
	\includegraphics[width=\linewidth]{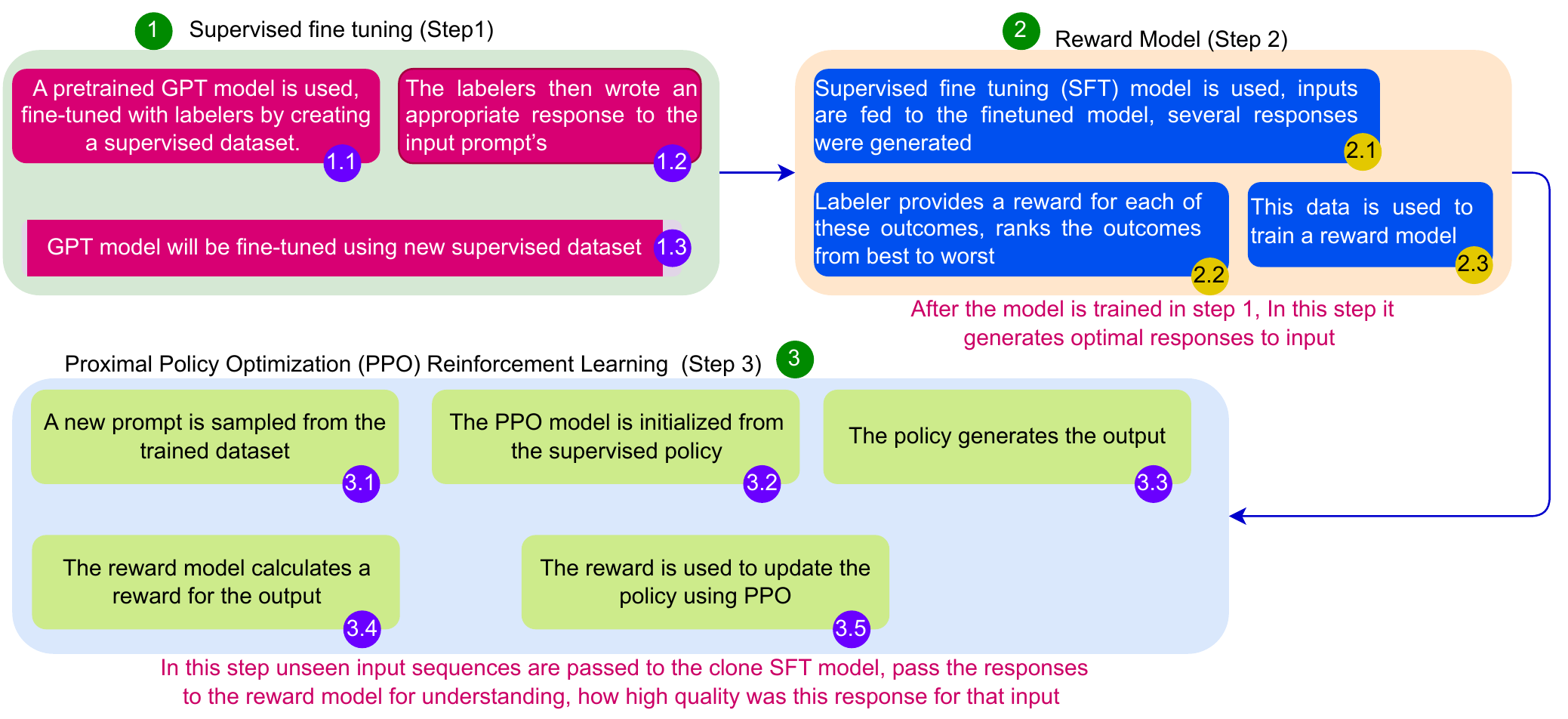}
	\caption{How does GPT Work.}
	\label{fig:working}
\end{figure*}

GPT models work by using a transformer which is a neural network architecture that processes the input sequences of natural language text \cite{work2}. The GPT model uses unsupervised learning techniques to pre-train this transformer architecture on a significant amount of text input \cite{hendy2023good}. The model gains the ability to anticipate the subsequent word in a sequence based on the preceding words during pre-training. Language modelling is the process that enables a model to discover the statistical connections between words and their context in training data. Fig. \ref{fig:working} shows the various stages of GPT operation. The first step entails supervised fine-tuning, the second step involves producing optimal responses to input, and the third step involves proximal policy optimization and reinforcement learning.

The model can be fine-tuned for particular tasks, like text classification or text production, after pre-training. The model is trained on a smaller dataset that is unique to the work at hand during fine-tuning, and the model's parameters are changed to maximize performance on that task \cite{zhang2023complete}. Fig. \ref{fig:transformer} shows the general transformer architecture of GPT.\\
When used for text creation, GPT models create text by anticipating the following word in a series based on the previously created words. Depending on how it has been modified, the model can produce text that is comparable to the input text or that adheres to a certain theme or style. Fig. \ref{fig:Inputtransformer} projects the GPT model's transformer architecture and input transformations for fine-tuning different tasks.
\subsection{ Comparisons of GPT Versions}

\begin{figure*}[!ht]
	\centering
	\includegraphics[width=\linewidth]{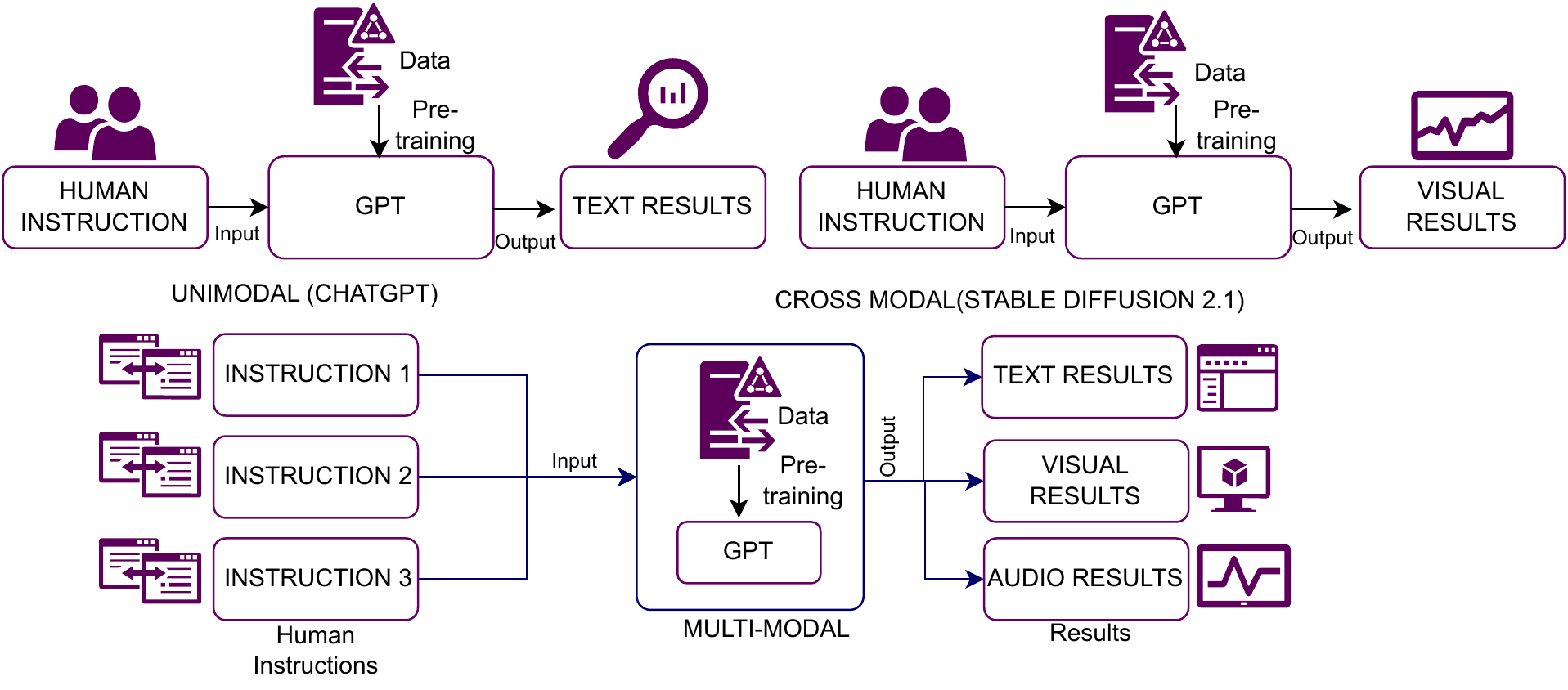}
	\caption{A comparison between unimodal, cross-modal, and multimodal modal GPTs.}
	\label{fig:Types}
\end{figure*}
There are several versions of GPT models each having their own features and capabilities. Table \ref{tab:comparisions} presents a comparison of various versions of the GPT models. The table presents the following details like year of release of the GPT model, parameters, tokens generated, input type, features of each model, drawbacks of each model, and the size of each model.\\
Generative AI (GAI) models are of different types like unimodal, cross-modal, and multimodal. The first type is unimodal which rely on a single type of input, such as text or images. The cross-modal, on the other hand, can process multiple types of inputs and relate them to each other. The Multimodal is the most complex type of AI as it can process and integrate information from multiple modalities, such as speech, text, images, and even physical interactions with the environment. GPT adopts only unimodal and multimodal types where ChatGPT is said to be unimodal, while GPT-4 is multimodal. Fig. \ref{fig:Types} is an illustration that distinguishes between unimodal, cross-modal, and multimodal Generative AI models.

Overall, GPT models have demonstrated outstanding performance with NLP, by enhancing each iteration  and its predecessor' capabilities. Each model, however, also has its own restrictions and drawbacks, such as restricted output control, lack of diverse data, and ethical concerns. While selecting a GPT model for a particular task, researchers and developers should carefully take these factors into account \cite{comparisions}.\\
In detail, this section describes the evolution, and architecture of GPT and compares the different versions and types of GPT.
\section{Enabling Technologies}
\begin{figure*}[!ht]
	\centering
	\includegraphics[width=\linewidth]{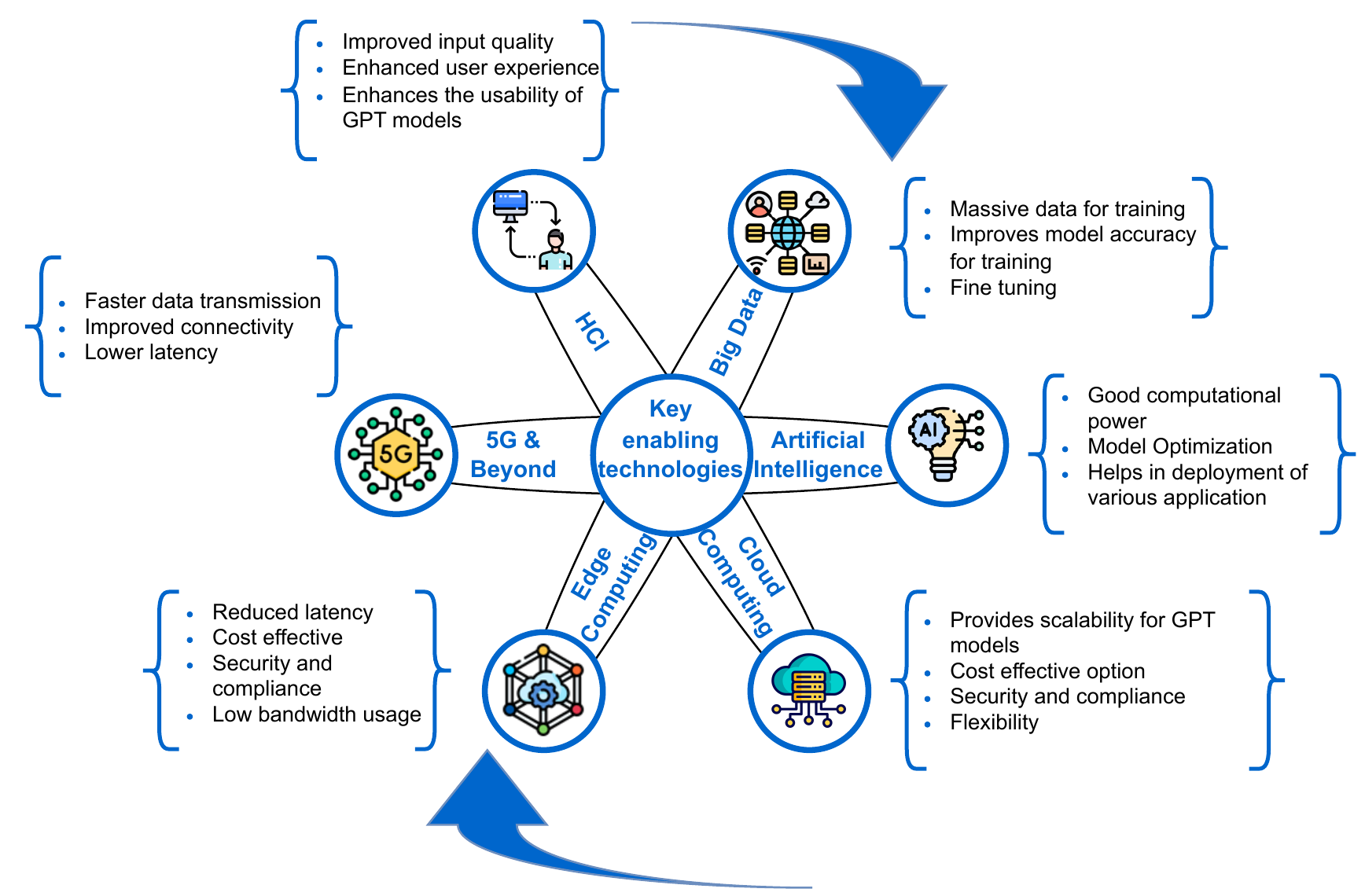}
	\caption{Enabling technologies of GPT models.}
	\label{fig:Technologies}
\end{figure*}

GPT is a convergence of several technologies. It is enabled by the latest technologies like Big data, AI, Cloud Computing, EC, 5G and beyond networks, and HCI. In this section, we provide an overview of enabling technologies related to GPT. The major technologies that constitute the GPT models are depicted in Fig. \ref{fig:Technologies}.

\subsection{Big Data}

Big data refers to the vast amounts of structured and unstructured data generated by businesses, individuals, and machines. The proliferation of new technologies, such as the IoT, has led to an explosion of data production from sources like social media, sensors, and transaction-based systems \cite{pramanik2023analysis}.\\

The emergence of big data has revolutionized the way organizations approach data analysis and decision-making. The training provided by this massive amount of data has yielded valuable insights for the use of advanced models like GPT in the field of NLP \cite{zaremba2023chatgpt}. The GPT models utilize DL and big data for natural language generation, with GPT-4 being the most advanced model to date \cite{aleksicdeep}.
The training data for GPT models typically include millions or even trillions of data from a diverse range of sources, such as books, articles, websites, and social media platforms. This large and diverse training data helps GPT models capture the variations in language usage, making them more accurate and effective at NLP tasks. As a result, GPT models may be used for a variety of tasks, including question-answering, text summarization, and language translation \cite{bubeck2023sparks}. Moreover, since GPT models can learn from a variety of data sources, they can be tuned for certain tasks and domains, making them very adaptive and versatile. GPT model has the potential to be utilized for a variety of activities, including the creation of images and videos in addition to its excellent language processing capabilities \cite{katz2023gpt}. \\

While big data presents numerous benefits to GPT, by enabling the models to get trained with large amounts of data, it also presents several challenges \cite{beerbaum2023generative}. GPT is trained on a variety of data, large amounts of data, and also sensitive data. Thus, ensuring data accuracy, privacy concerns, and ethical use of data are some of the challenges that must be considered. However, with the continuous growth of available data, GPT models will become even more advanced and capable of performing increasingly complex tasks \cite{dida2023chatgpt}. The future of big data as an enabling technology for GPT models is promising, with the potential to revolutionize the field of NLP. As technology continues to advance, organizations must prioritize ethical considerations and data accuracy to fully harness the benefits of big data and GPT models.
\subsection{Artificial Intelligence}

AI refers to the simulation of intelligent behaviour in machines that are programmed to learn from their experience to reason, understand natural language, and perceive their environment \cite{pereira2023systematic}. AI gives machines the ability to sense their surroundings, deal with what they see, handle issues, and take action to reach a particular objective. The importance and capability of AI is growing all the time. \\

AI enables GPT models to allow machines to comprehend and react to human language. There are several ways in which AI can continue to help improve GPT and make it more powerful and effective in its language generation capabilities \cite{trajtenberg2018artificial}.\\ 
The following are the several ways through which AI can make GPT models more powerful:
\begin{enumerate}
\item Fine tuning
\item Dialogue generation
\item Natural language understanding
\end{enumerate}
GPT's model performance on particular tasks can be enhanced by utilizing AI approaches. For instance, it can be trained on a large corpus of text from a particular field such as legal documents or medical literature to better grasp and produce language in that field \cite{lund2023chatting}. Considering dialogue generation, AI techniques such as reinforcement learning and sequence-to-sequence models can be used to enable GPT generate more natural and engaging dialogue in conversational contexts. Similarly, AI techniques such as semantic parsing and named entity recognition can be used to help GPT better understand the meaning of language and the relationships between words and phrases. This can enable it generate more accurate and coherent language \cite{haluza2023artificial}.\\
The development and enhancement of GPT model language production capabilities depend heavily on AI, and GPT's capabilities will continue to be growing by continuous research and development in AI.

As GPT models become more advanced, there are growing concerns about the potential for them in reinforcing biases and propagate harmful or offensive content \cite{tu2021limits}. Some of these concerns also include bias which can lead to unintended discrimination and unfairness, lack of understanding of the context that can lead to misunderstandings or incorrect responses, poor data quality can lead to inaccurate or biased models, ethical concerns like privacy and autonomy \cite{mathew2023artificial}. AI models like GPT require significant amounts of computational power to train and run, which can have a significant environmental impact due to their high energy consumption \cite{reed2021theology}.\\
Though AI has a great deal of promise, it's critical to be aware of the underlying issues and make efforts to fix them to ensure that it is utilized responsibly and morally for GPT.
\subsection{Cloud Computing}
Cloud computing refers to the on-demand availability of computer resources, such as storage, processing power, and applications, delivered over the internet \cite{akbar2023security}. The GPT model's successes are possible not only because of algorithmic evolution but also increased computational capabilities i.e. exponential growth in hardware (computational power, storage capacity), cloud computing, and related operational software \cite{gupta2023review}. The applications for cloud and EC working together such as natural language generation, image completion, or virtual simulations from wearable sensors see that the work is made more compute-intensive \cite{sajja2023platform}.\\
GPT models need a lot of computational power to analyze a lot of data, and cloud computing offers the scalability required to cope with demand spikes. Without worrying about the constraints of on-premises hardware, GPT models can rapidly and easily scale up or down as needed with cloud computing \cite{etro2009economic}. Cloud-based platforms like Amazon Web Services (AWS) or Google Cloud Platform (GCP) provide access to distributed computing resources that can be used to train GPT. Since cloud computing provides web-based solutions and thereby does not require the purchase and maintenance of costly hardware, it can be a cost-effective choice for a GPT model. By utilizing cloud computing, the GPT model can only pay for the computing resources it uses \cite{yamaoka2022experience}. The other added advantage of cloud computing in GPT is, it gives GPT models the freedom to access computing resources whenever it wants, from any location in the world. This makes GPT models more accessible to users by enabling smooth operation across a variety of gadgets and platforms \cite{ressmeyer2019deep}. Cloud computing providers offer high security and compliance standards, which can protect the GPT model and its data from online dangers. Cloud service providers also possess the knowledge and tools necessary to effectively address security problems and stop data leaks. Cloud-based storage services, such as Amazon S3 or Google Cloud Storage, provide scalable and reliable storage for GPT's data.\\

Despite the advantages of cloud computing where it can help GPT models to operate more efficiently, effectively, and securely, there are also a few technological aspects where it creates a drawback for GPT \cite{balkus2022improving}. To function properly, the GPT model needs a sizable amount of computing power and data storage. These resources can be accessible online with cloud computing. As a result, continued operation of the GPT model requires a robust and dependable internet connection, and any breakdown in connectivity may result in delays or even data loss. There are some security concerns when storing sensitive data, such as personal information or trade secrets, in the cloud which can be risky if proper security measures are not in place \cite{gupta2023review}. While cloud computing can be more cost-effective than building and maintaining an in-house computing infrastructure, it can still be expensive for long-term use. It also suffers issues like performance variability, limited availability etc., 
\subsection{Edge Computing}
The rapid growth of IoT, a large amount of data from several IoT devices, and also cloud services have necessitated the emergence of a concept called EC. EC is an open AI and distributed design with decentralized computational power. In EC, there is a lesser need for clients and servers to communicate over long distances, which lowers latency and bandwidth utilization.\\
Instead of depending on centralized data centers, EC entails bringing computing capacity and data storage closer to the consumer \cite{hua2023edge}.\\

 In GPT, where there is a need for real-time data analysis, EC plays a major role in faster processing and better efficiency in producing good results\cite{9783185}. GPT models are typically large and complex, requiring significant processing power to run. By deploying GPT models on the edge devices, closer to the source of data, latency can be reduced in replying to users who seek information through the GPT models by eliminating the need to move data back and forth from end devices to the cloud. Since EC maintains data near the periphery and away from centralized servers, it can offer improved security and more privacy protections in the case of the requests made by users through GPT \cite{yu2022measurements}. GPT models utilize a lot of data for learning and thereby the cost of data transfer also increases with data volume. EC can aid in controlling data transfer expenses. EC can also help in lowering the amount of bandwidth by pre-processing the data even before transferring it to the cloud. Particularly when analyzing photos or videos, GPT models can produce a lot of data \cite{yuanparatra}. EC accelerators, such as graphics processing units (GPUs) and field-programmable gate arrays (FPGAs), can be used to speed up GPT model inference and training. These accelerators can be integrated into edge devices or edge servers, providing more efficient processing of GPT models.\\
EC and GPT models make a great combination. Comparative to cloud data centres, edge devices may have constrained computation and storage capabilities \cite{zhou2022deep}. This might limit the scope of GPT models that can be installed on edge devices in terms of size and complexity. Since GPT models handle large and varied data, EC can also increase security risks and data privacy concerns. Implementing EC in existing infrastructure can be difficult and require significant investment in hardware, software, and networking components. This can be a barrier for many organizations which are using the GPT model and EC \cite{li2023ubinn}.

\subsection{5G and beyond networks}

5G networks represent the latest generation of cellular networks that promise faster data speeds, lower latency, and the ability to connect a vast number of devices simultaneously \cite{benzaid2021trust}. 5G and beyond networks enable faster data transmission speeds than previous generations of cellular networks, which can help in training and deploying larger and more complex language models. This can result in faster training times and better performance. 5G and beyond networks can provide lower latency than previous generations of cellular networks, which can reduce the time required for communication between GPT and other devices, such as servers or other language models \cite{8412589}. This can improve the real-time response of the GPT model for applications that require quick and accurate language processing. 5G and beyond networks offer improved connectivity options, such as increased capacity and more reliable connections, which can help in scaling up the deployment of the GPT model for large-scale language processing tasks. With the deployment of 5G and beyond networks, EC is becoming more prevalent. This means that a GPT model can potentially be deployed closer to the end-user, reducing the latency and improving the response time for applications that require real-time language processing \cite{zhang2023towards}. Ultra-Reliable Low-Latency Communication (URLLC) is a key feature of 5G networks. In the context of GPT language models, URLLC can enable real-time and reliable communication between multiple devices, such as edge devices, cloud servers, and end-users \cite{8491078}.\\

Though 5G and beyond technology offers potential advantages to GPT models, it is also important to note that the actual impacts of this technology may change depending on how it's implemented and used. 5G enables the access to uncontrolled access to the Internet,it may attract cybersecurity risks and privacy concerns \cite{gonzalez2019road}. Also, as GPT uses a large amount of data for analysis it could also cause privacy concerns. 5G and beyond networks in GPT models need high infrastructural requirements which is a costly process.
\subsection{Human Computer Interaction}

HCI, which is multi-faceted, concentrates on the design of computer technology and, in particular, on how people and computers communicate with each other \cite{rogers2023interaction}.\\

HCI has a greater influence over GPT models. As a language model, GPT is designed to interact with humans by generating natural language responses to input text. HCI research can help designers create more effective input mechanisms for the GPT model, such as natural language interfaces, that allow users to communicate more easily and accurately with the model \cite{liu2023creative}.
HCI also helps in enhancing the GPT model's user experience by creating interfaces that are more intuitive and user-friendly. This makes it easy for the users to interact with GPT models and understand their responses \cite{hamalainen2023evaluating}. HCI also estimates the performance of GPT models by evaluating their responses with real-time users and identifies the areas where the model needs improvement, thereby improving its reliability and accuracy. HCI enhances the usability of GPT models by reducing the time and effort required for the users to interact with \cite{shafeeg2023voice}.\\

While HCI can be incredibly helpful in improving the design and usability of GPT models, there are also some potential drawbacks to consider. If the research is not conducted with a diverse group of users, HCI can introduce biases into the design of the GPT model. HCI techniques can be expensive and time-consuming. As GPT models become more complex, it may become more difficult to design interfaces and input mechanisms that are both effective and user-friendly \cite{zhang2023hivegpt}. HCI may not always be able to provide the necessary insights or feedback to drive improvements in GPT models. There are also ethical concerns around the use of GPT models, including issues related to privacy, bias, and the potential misuse of the technology \cite{nori2023capabilities}.  As GPT models become more complex, it may become more difficult to design interfaces and input mechanisms that are both effective and user-friendly.
\section{Impact of GPT models on various Applications}
\begin{figure*}[!ht]
	\centering
	\includegraphics[width=\linewidth]{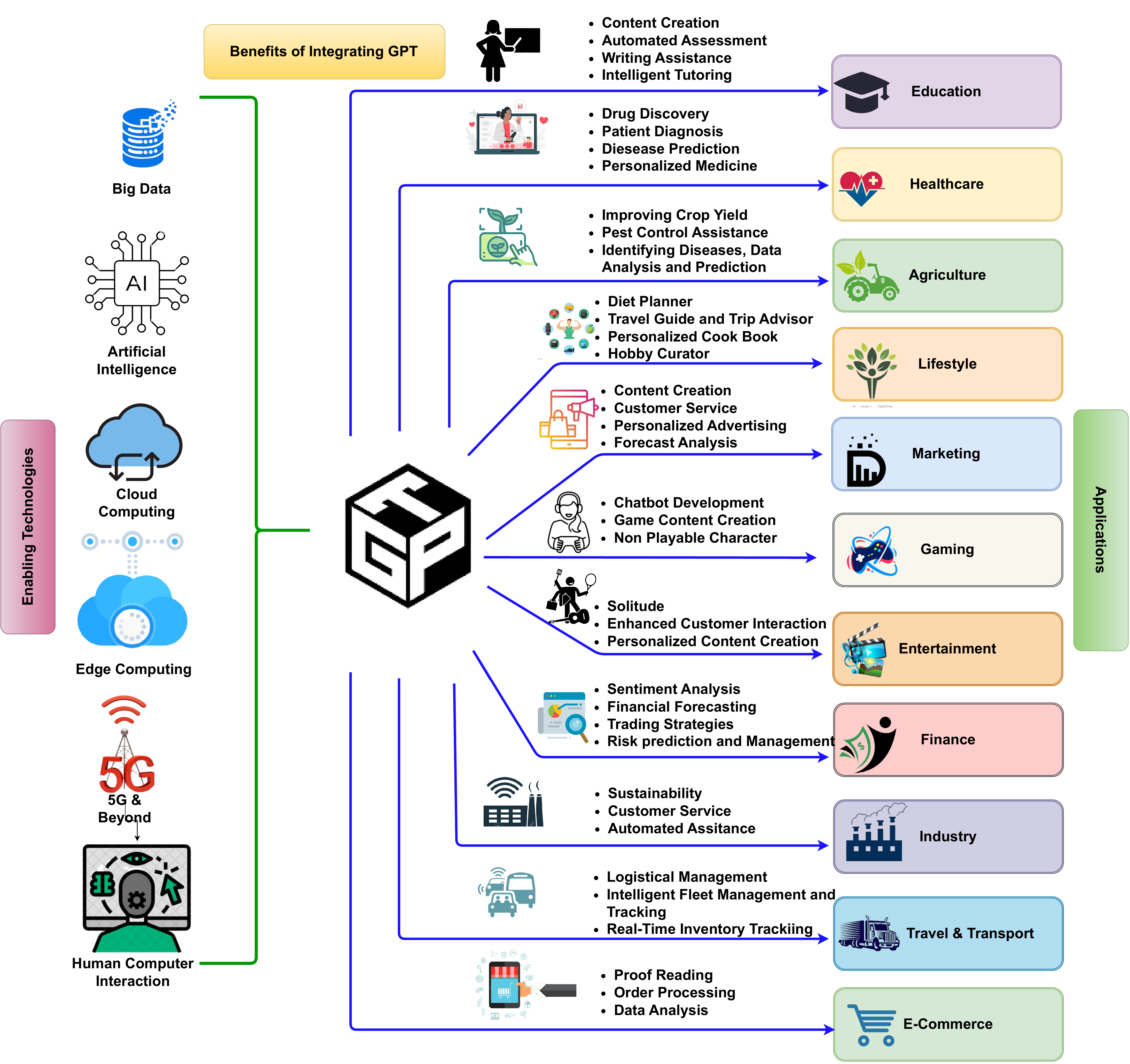}
	\caption{The impact of GPT models on various applications.}
	\label{fig:Apps}
\end{figure*}

GPTs have made significant progress, and its impact is being felt across various industries 
 like education, healthcare, industry, agriculture, travel and transport, e-commerce, entertainment, lifestyle, gaming, marketing, and finance. This section provides valuable insights on the impact of the GPT models in the aforementioned applications as depicted in Fig. \ref{fig:Apps}.

\subsection{Education\hl{}}
\subsubsection{Introduction}
Education has been around for centuries, with traditional education being the most common form. Traditional education involves a teacher imparting knowledge to a group of students in a physical classroom. While successful, traditional education can be restrictive and inflexible, limiting students' ability to learn at their own pace and in their preferred style. It can also be limited by geography, as students need to be physically present in a classroom to learn. Technology has emerged as a solution to some of these issues, allowing for personalized learning experiences and more engaging, accessible resources. Online learning platforms, digital textbooks, and multimedia tools offer students access to a vast array of resources from anywhere in the world. Technology can also facilitate collaboration and communication among students and teachers, leading to a more dynamic and interactive learning experience. Distance learning, hybrid learning models, and online classes are examples of how technology can help break down the barriers of traditional education, making learning more flexible, efficient, and effective. By integrating technology into traditional education, we can create a more personalized and effective learning experience, benefiting students worldwide.
\subsubsection{Impact of GPT in Education}
The field of education is constantly evolving, with advancements in technology playing a significant role in shaping the way we learn and teach. One such technology that has the potential to transform the education industry is GPT. As a large language model trained on a vast amount of data, GPT can generate human-like text that is coherent and informative, making it a valuable tool in developing educational content such as textbooks, study guides, and course materials. Furthermore, GPT can be used to analyze and summarize complex text, which can help educators and students save time and increase comprehension. With its ability to support NLP applications and create intelligent tutoring systems, GPT has the potential to revolutionize the way we learn and teach. In this context, following section will explore the different ways in which GPT can contribute to the education industry and transform the future of learning.

\begin{itemize}
    \item {Intelligent Tutoring:} Intelligent tutoring is a teaching approach that uses AI and ML to provide personalized and adaptive instruction. It analyzes student performance data, understands their strengths and weaknesses, and generates customized learning paths. It provides immediate feedback, personalized guidance, and remedial support. It is effective in improving learning outcomes, increasing student engagement, and reducing learning time. With advanced natural language processing capabilities, GPT can enhance the personalized and adaptive instruction provided by intelligent tutoring systems. It can analyze natural language input from students, enabling intelligent tutoring systems to better understand and respond to their queries, needs, and preferences. It can also generate personalized feedback and assessment based on the individual learning progress of each student, helping them to identify and address their knowledge gaps and improve their performance. GPT can also analyze student performance data and generate adaptive learning paths that provide customized instruction and remediation, ensuring that each student learns at their own pace and achieves their learning objectives. Additionally, it can create interactive dialogue systems that simulate natural conversations between students and virtual tutors, making learning more engaging, interactive, and personalized \cite{tack2022ai}. The authors in \cite{nori2023capabilities} have identified that GPT-4 model outperforms general-purpose GPT-3.5 model as well as GPTs (Med-PaLM, a prompt-tuned version of Flan-PaLM 540B) specailly trained on medical data. The authors have tested GPT-4 models' ability to explain medical reasoning, personalize explanations to students, and interactively craft new counterfactual scenarios around a medical case.

    \item {Learning assistance and material development:} Learning materials are critical in education as they provide a structured way for students to acquire knowledge and skills. They can be tailored to meet the needs of diverse learners and make learning more engaging and effective, supporting teachers to create a more dynamic and interactive learning environment. GPT can contribute to creating learning materials by automating content generation, providing multilingual content creation, language correction, personalized content creation, conducting topic research, and generating assessments. It saves time and effort for educators and publishers, improves the accuracy and readability of material, and makes learning more engaging and effective. GPT can generate high-quality content such as summaries, quizzes, and lesson plans based on specific learning objectives, making learning accessible to a wider audience. It can analyze written content and provide suggestions to improve grammar, punctuation, and readability. GPT can also assist in research writing by suggesting ideas for structure, rephrasing and organizing content, and identifying gaps in research \cite{lecler2023revolutionizing}. Moreover, GPT can also provide personalized feedback based on individual learning progress, enhancing the development of more comprehensive and informative learning materials.
     
    \item {Automated Assessments:} 
    Automated assessment in education uses technology to evaluate students' learning outcomes, providing immediate feedback and reducing potential bias in grading. It can also help teachers identify areas where students may need additional support, enabling them to tailor their teaching methods to better meet individual needs. GPT with its advanced natural language processing skills, can help in automated assessment by analyzing and grading student responses to various types of assessment questions, including essays and short answer questions. It can also provide feedback to students \cite{baidoo2023education}, such as highlighting areas for improvement and suggesting further reading or resources. GPT's natural language processing capabilities can help to identify the meaning and context of students' responses, making automated assessment more accurate and effective. Additionally, GPT can generate personalized learning materials and exercises based on students' assessment results, supporting educators to create more tailored and effective learning experiences. The authors in \cite{o2023system} have used Chat GPT in evaluating the students' assignments such as quiz style questions, and also in generating relevant practice problems to improve content retention and understanding. The results were promising in the classroom. The authors believe that Chat GPT has the significant ability in reducing the load of instructor without compromising students' learning outcomes.
    
    \item {Fostering Creativity:} Creativity thinking plays a vital role in education by encouraging students to think beyond traditional boundaries and develop innovative solutions to complex problems. It helps students to approach learning with an open mind and a willingness to explore new ideas, leading to greater engagement and motivation. GPT's ability to generate human-like responses and creative writing can aid in improving creativity. It can help improve creativity by generating new and innovative ideas based on vast amounts of data and information. By analyzing patterns in language and identifying connections between different concepts, GPT can suggest novel approaches to teaching and learning. Additionally, GPT can also generate creative prompts or challenges for students, encouraging them to think outside the box and approach problems in unique ways \cite{alam2020possibilities}. GPT can also analyze and evaluate students' creative work, providing feedback and suggestions for improvement. So, GPT can be a valuable tool for promoting and enhancing creativity among students and faculty members.

\end{itemize}

\subsubsection{Challenges}

There are several advantages to incorporating GPTs in education, but it is essential to acknowledge the potential limitations. While GPTs can quickly generate information, they may impede students' critical thinking and problem-solving skills. Furthermore, learners who benefit from personal interaction with instructors may find the lack of human involvement disadvantageous. GPTs rely on statistical patterns, so they cannot provide a comprehensive understanding of the material being taught \cite{baidoo2023education}. Privacy concerns arise when using sensitive student data in GPTs for educational purposes. Additionally, since GPTs cannot provide citations, it is challenging to identify the source of information generated. The cost of maintaining GPT may be prohibitive for schools and educational institutions with limited resources. Finally, distinguishing between reliable and unreliable information generated by GPTs can be difficult, so it is necessary to have human oversight to ensure data accuracy and regulate access.
 
\subsubsection{Summary}
GPT offers numerous advantages in the education sector, including personalized and adaptive instruction, automated assessment, creative writing support, and research writing assistance. They have the potential to revolutionize teaching by creating lesson plans and activities, responding to natural language queries, and integrating multiple digital applications. However, there are also challenges to consider, such as the potential negative impact on critical thinking and problem-solving skills, lack of human interaction, data security and privacy concerns, inability to provide full comprehension, lack of citations or sources, high cost of maintenance, and potential for producing unreliable information. Further research is needed to explore human-computer interaction and user interface design to integrate GPT into educational workflows while ensuring that the information they provide is accurate and reliable.
\subsection{Healthcare}
\subsubsection{Introduction}
Before technology became widespread in healthcare, healthcare services were primarily delivered through face-to-face interactions between healthcare professionals and patients. Traditional healthcare faced several challenges, including limited medical instruments, paper-based health records, patients receiving care mostly in hospitals or clinics, physical travel requirements to receive medical attention, and limited medical research. Despite these challenges, traditional healthcare still provided valuable medical services to patients. However, with the introduction of technology, healthcare has become more efficient, accessible, and personalized, resulting in improved patient outcomes and better overall healthcare services. Technology has become an essential aspect of society, as reflected in the significant investments made in this sector. Despite the advancements in technology, the healthcare industry still faces various new challenges, including access to healthcare, high costs, personalized medicine, data privacy and security concerns, and an aging population. However, technology has the potential to address these challenges and improve the efficiency of healthcare services.
\subsubsection{Impact of GPT in healthcare}
Recent years have seen significant advancements in technology, including in the healthcare industry. Biotechnology, medical devices, and pharmaceuticals have undergone transformations through the use of cutting-edge technologies like DL \cite{chen2018rise} and ML \cite{pillai2022machine}. Currently, the healthcare sector is utilizing various forms of AI techniques for medical research and providing medical services. One such technique is the GPT features of NLP, which hold immense potential for the healthcare industry. GPT can help to overcome several challenges in healthcare in multiple ways. For instance, it can be used to develop intelligent systems that assist doctors in making accurate diagnoses and providing clinical assistance \cite{kim2022pre} \cite{kalyan2022ammu}. GPT can also analyze large volumes of medical data and generate reports. Furthermore, it has potential applications in drug discovery \cite{liu2021ai} \cite{vamathevan2019applications}, personalized medicine, patient diagnosis, medical image analysis, analyzing electronic health records, clinical decision support systems, and disease prediction.
\begin{itemize}
    \item {Drug Discovery:} Recent AI and machine learning techniques \cite{gawehn2016deep} \cite{lavecchia2019deep} are having the potential to contribute to the growth and development of drug discovery. GPTs are capable of learning new patterns and relationships \cite{urbina2022dual} in the dataset they were trained on. This capability can be used in drug discovery to aid in the identification and design of potential new drugs with desired properties\cite{segler2018generating}. One of the key challenges in drug discovery is finding compounds that can interact with specific parts of the body. GPT can help in this process by learning the patterns and relationships from large databases of known compounds \cite{liu2021ai}. GPT can be trained on large sets of chemical databases to analyze chemical reactions and their outcomes. This can help suggest potential combinations of new drugs using the analyzed data. These new drugs can also be analyzed using GPT to test their efficacy and toxicity. 
   
    \item {Diagnosis:}
    GPT can be used in medical diagnosis by analyzing patient data. It can help to analyze medical records and extract information such as patient demographics, symptoms, and medical history. This can help medical professionals provide effective patient care and improve outcomes. The recent release of GPT-4 has the ability to support multimodal information, allowing it to analyze images as input and produce text results as output \cite{ahsan2023chatgpt}. It is recommended to use AI systems such as a GPT, as clinical support tools to assist medical professionals in diagnosing and treating patients, but they should not be relied upon as the sole source of medical advice or decision-making. GPT can also be used to identify rare diseases by analyzing patient's complete information. The authors in \cite{levine2023diagnostic} have used a general-purpose GPT based on GPT-3 model for patient diagnosis and triage. The model has given a triage accuracy of 70\% which was worse than a physician. But, in next subsequent weeks, the accuracy has improved to 92\% which is close to the performance of a physician. In diagnosis, GPT-3 model has given 88\% accuracy. For emergency cases, GPT-3 has given 75\% accuracy whereas physician has given 94\%.

 \item {Disease prediction:} GPT has great potential in disease prediction \cite{alghanmi2021probing}. By analyzing large amounts of medical data, including patient records, medical images, and clinical trials, these pre-trained language models can learn patterns and make predictions about the likelihood of a patient developing a particular disease. For instance, trained healthcare GPTs can be used to predict the occurrence of diseases such as diabetes, heart disease, and cancer by analyzing various parameters, including the patient’s medical history, age, family history, and lifestyle. It can also be used to predict the likelihood of a rare disease This helps in the early detection of high-risk patients so that medical personnel can take necessary measures and suitable medicines to reduce the risk of developing the disease. The medical practitioner and author in \cite{ali2023potential} have recommended using GPT-4 models’ ability of NLP in bariatric surgery.
    
    \item {Personalized medicine:}
   The COVID-19 pandemic has highlighted that not all body systems are clinically similar. For instance, during the pandemic, medicines like Remdesivir and Tocilizumab have been effective for one category of patients but do not affect another category of patients with similar clinical metrics, as they progress from a mild or moderate level of infection to a severe stage \cite{vulturar2022therapeutic}. This highlights the need for personalized medicine in today's world. GPT can be used to identify variable patterns of data to predict or classify hidden or unseen patterns, which can be used for exploratory data analysis. GPT provide the possibility of identifying personalized medicines \cite{wang2021pre} based on the clinical, genomic, and nutritional data of patients. The dietician and the author in \cite{arslan2023exploring} have observed that the utilization of Chat GPTs has significantly decreased obesity rates among patients by offering personalized recommendations regarding nutrition plans, exercise programs, and psychological support. This approach allows for the development of customized treatment plans that cater to the specific needs of individuals, leading to a more efficient method of treating obesity with the assistance of Chat GPT.
\end{itemize}
\subsubsection{Challenges}
While GPT is a powerful language model with numerous applications in healthcare, it is not without its challenges. The primary challenge is data bias. As GPT models are also learning models, the significant drawback of biasing is also applicable to GPT. GPT can be susceptible to bias. If the data used to train the model is biased, the model will learn from it and replicate the bias. This leads to incorrect treatment and predictions. Another challenge is the transparency of the model. GPT is complex to understand and interpret. This lack of transparency in technology can make doctors and medical personnel not believe in the predictions, which may result in a hesitancy to trust and adopt technology \cite{blanchard2022ethical}. Another important concern is security and privacy issues. As it is a model to be trained on data, there is a huge amount of sensitive information about the patients to be used to improve the algorithm and its performance. This results in significant security and privacy concerns related to the use of GPT in healthcare. The final and important challenge is limited clinical validation. GPT are showing promising improvement in various fields of healthcare, such as drug discovery, and disease prediction. But still, their effectiveness and accuracy in medical research and clinical settings have yet to be validated. More research and clinical trials are required to prove that GPT can transform the medical industry with full trust.
\subsubsection{Summary}
GPT have the potential to revolutionize the healthcare industry by contributing to drug discovery, personalized medicine, clinical support in making decisions, diagnosis support, and disease prediction. This can be helpful for human beings to predict the disease in advance and treat it through proper medicine. However, there are significant challenges that are to be addressed, such as technology adoption, data bias, regulatory challenges, and security and privacy issues. It is so important to analyze and evaluate the benefits and risks of using GPT in healthcare and to continue to monitor their development and implementation.
\subsection{Industry}
\subsubsection{Introduction}
An important economic transition from agriculture and handicrafts to large-scale industry and automated production was achieved by the industrial revolution. Efficiency and productivity were raised as a result of modern equipment, energy sources, and labour arrangements. New opportunities and challenges have been created as a result of the quick development of new technologies in both the workplace and other industries \cite{balsmeier2019time}. The utilization of big data is a well-known technology-driven trend. Nowadays, companies have access to enormous volumes of data that may be examined to uncover insightful information. Big data can help businesses make wise decisions and discover areas for development. AI is another innovation that is changing industries. AI systems have the ability to analyse complex data, automate procedures, and make wise conclusions \cite{pesapane2018artificial}. This improves production by increasing its dependability, adaptability, and efficiency. The process of "digitalization," which includes incorporating digital technologies into every element of business, is creating industries to become more flexible, efficient, and valuable. Businesses may automate tedious work, improve client experiences, and streamline operations by implementing digital solutions. In today's digitally-driven world, adopting technological advancements is essential for maintaining competitiveness and promoting growth.

\subsubsection{Impact of GPT in Industry}
In industrial scenarios, GPT has the potential to be applied as a sustainability tool, assisting businesses in evaluating and enhancing their sustainability goals. Companies can improve supply chain tracking and query response by integrating pre-trained transformer models like ChatGPT with supply chain management platforms\cite{carvalho2023chatgpt}. Additionally, GPTs can offer modifications to the production process that might increase efficiency \cite{rathore2021towards}. GPT can also help users make knowledgeable decisions about how to use resources, allowing businesses to remain competitive while reducing their environmental effect. For example, the GPT-2 model has demonstrated efficacy in sentiment analysis, providing insightful data for numerous applications \cite{shah2023ensemble}.

\begin{itemize}
    \item  {Hospitality sector}
In the hospitality industry, hotels place a high focus on providing satisfying guest experiences. To ensure that every tourist is satisfied during their stay, this necessitates adapting to their requirements and preferences. Hotels may improve the guest experience in a number of ways by integrating GPT into their website or mobile application. Hotels may respond to consumer inquiries in a timely and precise manner by utilizing GPT \cite{salminen2022creating}. Customers do not have to wait for human assistance when looking up information about facilities, booking procedures, or room availability. Customers' overall satisfaction with the hotel's services is increased as a result of the large reduction in client wait times. GPTs can also make it easier for visitors who speak multiple languages to communicate \cite{george2023review}. Hotels can offer a more inclusive and welcoming experience for visitors from other countries by removing linguistic obstacles. Hotels may provide their visitors with immersive and engaging experiences by combining GPT with AR technologies. For instance, customers can use their mobile devices to get AR guided tours of the hotel or nearby attractions, offering a distinctive and entertaining way to explore the surroundings and learn more about the hotel's amenities.GPTs integration into various aspects of the hospitality industry gives hotels the ability to deliver streamlined, tailored, and effective services, increasing client happiness and loyalty.
\item  {Fashion:} By providing highly customized user recommendations based on personal style, brand preferences, and particular clothing or accessory demands, collaborative filtering and AI algorithms have undoubtedly revolutionized the fashion business. The amount of personalization has been further increased in this context by the incorporation of GPT, dramatically altering the purchasing experience for customers \cite{el2023sentiment}. Fashion platforms may analyse a significant quantity of user data, such as browsing history, purchasing behaviour, and style preferences, using the advanced capabilities of GPT to produce tailored recommendations. Fashion platforms can direct consumers towards clothing options that fit their desire for sustainable fashion by including eco-friendly fabric selections into the system. GPT improve users' general fashion knowledge and confidence while enabling users to keep up with the most recent trends. The image-text retrieval skills of GPT significantly improve visual search capability in fashion platforms \cite{goei2021tackling}. Users may make more confident shopping decisions and minimize the need for returns by visualizing how various clothing items and accessories would appear on them without physically trying them on. The model may recommend the proper size for various brands and apparel products by taking into account a user's measurements, preferred fit styles, and historical data. The overall purchasing experience is enhanced and the frustration of wrong size is decreased. 
\item {Sustainability:}
Sustainable development means addressing current demands without sacrificing the capacity of future generations to address their own needs. Goals for sustainable development can be attained by implementing GPTs in a variety of sectors, including manufacturing and corporate operations \cite{bouschery2023augmenting}. The models can estimate where energy saving measures would be most useful by analyzing past data and patterns to provide insights into energy usage, pinpoint problem areas, and recommend opportunities for improvement. GPTs can aid in identifying sustainability-related problems, creating plans and strategies to solve them, investigating brand-new sustainable activities, keeping track of advancements, and conducting routine reviews. Companies can choose activities that will have the biggest positive impact by grading tasks and actions according to their impact on sustainability \cite{bommasani2021opportunities}.
The models can optimize supply chains for decreased carbon emissions, minimized waste, and improved resource efficiency by assessing elements including transportation routes, packaging materials, and supplier practises \cite{cowls2021ai}. This results in more environmentally friendly production, distribution, and sourcing procedures.
\end{itemize}
  
\subsubsection{Challenges}
There are many different industrial fields where GPT models can be applied; the three areas mentioned above are only a few. However, for optimal use, the industrial sector needs to be ready to adapt to a constantly changing environment. Public and corporate policies must be developed over the long term to promote the use of sustainable production techniques. For enterprises, deploying pre-trained GPT models can be a costly task. Continuous development and training are also required to accommodate new and evolving inquiries as client expectations change. Companies have to carefully assess the benefits and costs before implementing the GPT model because these continuing efforts raise the deployment cost \cite{benbya2020artificial}. For industries to fully benefit from GPT models, it is crucial to address issues with interpretability, data reliance, and ethical considerations. Industry may therefore take advantage of these GPT models' advantages, make wise decisions, and promote sustainable development.

\subsubsection{Summary}
GPTs have the ability to have a positive impact on society and business operations. They can speed up operations like accounting, sales, and marketing, increasing productivity. But before they are widely used, ethical problems need to be fully investigated. Technology products will change as GPT models develop. To reap the benefits and reduce dangers, it is essential to solve interpretability and data concerns. GPTs can have a tremendous positive impact on businesses, society, and the economy when they are used responsibly.
\subsection{Agriculture\hl{}}
\subsubsection{Introduction}
Traditional agriculture, a time-honored practice passed down through generations, sustains civilizations with its crop cultivation and livestock rearing methods. Rooted in a deep connection to nature, it emphasizes sustainability and local ecosystem understanding. Beyond providing sustenance and livelihoods, traditional agriculture preserves cultural heritage. However, it also faces challenges such as labor-intensive processes and shortages, inefficient resource utilization, vulnerability to pests and diseases, and limited access to real-time data and environmental impact. Today, by merging tradition with modernity, we have the opportunity to leverage technological advancements to enhance productivity, sustainability, and resilience while honoring the profound legacy of traditional agriculture for future generations.
\subsubsection{Impact of GPT in Agriculture }
GPTs have the ability to overcome the challenges of agriculture. It offers valuable advantages to the agriculture sector. It acts as a comprehensive knowledge source, providing information on crop cultivation, pest management, and soil health. By analyzing real-time data, GPT assists farmers in making informed decisions regarding optimal planting times and resource allocation. It plays a crucial role in identifying and addressing crop diseases and pests accurately. Moreover, GPT enables precision farming practices by utilizing sensor data and satellite imagery, ensuring precise irrigation, fertilization, and pest control. Additionally, it provides market analysis and price prediction, empowering farmers to navigate market conditions and optimize pricing strategies. GPT also supports farm management and planning, optimizing crop rotation and resource usage. By facilitating agricultural research and innovation, GPT contributes to advancements in crop breeding and sustainable practices. Embracing GPT in agriculture enhances decision-making, efficiency, and sustainability, ultimately promoting improved productivity and food security. For instance, GPT-4 can educate farmers about new methods and goods and warn them of potential issues or possibilities by analyzing data from many sources \cite{liu2023summary}.
\begin{itemize}
    \item {Improving Crop Yields:}

With its data analysis capabilities and real-time recommendations, GPTs plays a crucial role in enhancing crop yields. By examining historical yield data, weather patterns, soil conditions, and crop management practices, GPT identifies valuable patterns and correlations, providing insights and suggestions for optimal crop management techniques\cite{biswas2023importance}. It enables precision farming by integrating data from sensors, satellites, and IoT devices, granting timely guidance on resource allocation for improved efficiency. Additionally, GPT aids in the early identification and management of crop diseases and pests, minimizing yield losses through precise and prompt recommendations. Moreover, GPT supports crop breeding and genetic optimization by analyzing genetic data and plant characteristics, expediting the development of high-yielding and resilient crop varieties. Therefore, GPTs data analysis and decision support capabilities significantly contribute to enhancing crop yields and maximizing agricultural productivity\cite{farooq2019survey}.

    \item {Pest Control:}

    GPT offers significant support in the realm of pest control in agriculture. By analyzing extensive data on pests, including their behavior, life cycles, and characteristics, GPT can provide valuable insights for effective control measures. It aids in early pest detection by analyzing sensor data and satellite imagery, enabling proactive interventions to prevent pest spread and minimize damage \cite{wang2022residual}. GPT also assists in determining suitable pest control methods tailored to specific crops and pests, considering factors like environmental impact and sustainability. Additionally, it contributes to precision pest control by leveraging real-time data to optimize timing and dosage of interventions, reducing chemical usage and resistance risks. It also aids in identifying natural enemies and beneficial organisms, promoting natural pest control mechanisms such as habitat diversification and companion planting. Through GPT's data analysis and recommendation capabilities, it empowers farmers with informed decisions, leading to more effective and sustainable pest management strategies, ultimately reducing crop losses and enhancing agricultural productivity.

    \item {Identifying Diseases and Soil analysis:}

    GPTs offer valuable assistance in disease identification and soil analysis within the field of agriculture. With its ability to analyze extensive data sets, GPT can accurately identify crop diseases by processing information such as symptoms, historical data, and disease patterns. This enables timely and effective disease management strategies\cite{biswas2023importance}. Additionally, It plays a significant role in soil analysis by analyzing diverse soil-related data, including nutrient levels, pH, organic matter content, and soil composition. By interpreting this data, It provides insights into soil health and fertility, empowering farmers to make informed decisions regarding nutrient management, soil amendments, and cultivation practices. Moreover, GPT can identify complex interactions between soil conditions and crop diseases, helping farmers understand the relationship and take preventive measures accordingly. It also supports precision agriculture practices by integrating sensor data and satellite imagery to assess soil variations across fields, allowing for site-specific management strategies and optimized resource allocation. Furthermore, it also facilitates knowledge sharing and collaboration by analyzing and disseminating research findings, best practices, and disease outbreak information among agricultural communities. This collective intelligence enhances disease monitoring and control efforts on a broader scale.

\end{itemize}

\subsubsection{Challenges}
While GPT, provides significant benefits to agriculture, there are challenges to its implementation. GPT's effectiveness depends on the availability and quality of data, making insufficient or biased data a limitation. The interpretability of GPT's decision-making process is challenging due to its black-box nature, hindering trust and understanding. GPT's computational requirements and infrastructure can be demanding, posing difficulties for resource-constrained farmers. Language and domain-specific nuances can affect its performance, impacting accuracy and relevance. Ethical considerations surrounding data privacy and ownership need careful attention to ensure responsible use. By addressing these challenges, researchers and practitioners can unlock GPT's potential while ensuring its practicality and ethical implementation in agriculture.
\subsubsection{Summary}
GPT holds immense potential in agriculture, offering numerous benefits alongside notable challenges. Its data analysis capabilities empower farmers with informed decision-making in disease identification, soil analysis, and precision farming, leading to improved crop yields and sustainable practices. However, the effectiveness of GPT relies on data availability and quality, while its interpretability remains a challenge due to its black-box nature. Additionally, computational requirements, language nuances, and ethical considerations require careful attention. By addressing these challenges, the agricultural sector can harness the full potential of GPT, paving the way for more productive, efficient, and responsible farming practices.
\subsection{Travel and Transport\hl{}}
\subsubsection{Introduction}
Historically, animals have been used by people as their main source of transportation. But as the world's population increased, the demand for more effective transportation systems increased. Transportation-related technological advancements have fundamentally changed the sector in several ways. Business operations like order tracking, freight management, and customer support can be streamlined by automation employing AI-driven technologies. Companies can enable their employees to concentrate on more beneficial and profitable duties by automating these tasks\cite{dwivedi2023so}. With better transportation networks and logistics management systems that optimize routes and reduce transit times, technological developments also enable speedier delivery times. In terms of product development, technical advancement has paved the way for the development of innovative vehicles, infrastructure, and logistics systems, leading to the production of more sophisticated and effective transportation choices. Another noteworthy benefit of technology advancement in logistics and transportation is increased customer service. Inquiries and problems can be handled quickly and efficiently by chatbots and customer support systems powered by AI, improving the entire customer experience\cite{haluza2023artificial}.
\subsubsection{Impact of GPT in Travel and Transport }
Companies can learn about customer preferences in real time by using GPTs in logistics and transportation, which results in better personalization and more customer satisfaction. GPTs leverage NLP approaches to interpret customer requirements and preferences, enabling customized suggestions as well as guidance in the logistics and transportation processes. The most effective routes and forms of transportation can be recommended using GPTs, which can analyse a large amount of data, including traffic patterns, weather conditions, and delivery requirements\cite{biswas2023prospective}. In addition, GPTs can be used as travel planners, allowing visitors to enter their travel budget, duration, and destination to create customized itineraries. For travel agencies, this personalized approach increases consumer satisfaction and revenue.
\begin{itemize}
    \item {Logistical Management:} GPTs can be quite important in the context of shipping logistics. They can automate the creation of shipping labels, eliminating up manual entry and lowering the possibility of mistakes. Additionally, GPTs can have access to real-time tracking data and can integrate GPS data and sensors to provide businesses and customers with precise and up-to-date shipment status information. Companies can successfully monitor shipments with the use of GPTs, geographic information systems (GIS), and routing algorithms\cite{haluza2023artificial}. Organizations can track shipments in real-time and ensure visibility throughout the supply chain by utilizing GPS data and sensor technology\cite{helo2020real}. Customers can receive precise updates on their shipments using this real-time information, which will improve their experience overall. Overall, the use of GPTs into shipping logistics results in increased automation, efficiency, and client satisfaction. 
    \item {Intelligent Fleet Management and Tracking:} Companies can get real-time fleet updates by utilizing GPT models, which enables them to track vehicles quickly and precisely. GPT models' underlying technology also supports proactive fleet management. GPTs can identify possible problems or maintenance needs before they develop into expensive breakdowns or accidents by analyzing data from a variety of sources \cite{kadel2022emergence}. With this knowledge, companies may take preventative measures, such as planning maintenance or quickly fixing developing problems, ultimately saving time and money by preventing unintended delays. Additionally, GPTs can provide clever alerts and notifications. Businesses can receive alerts when vehicles arrive at specified areas by setting up specific triggers, which enables better coordination and customer service \cite{benbya2021special}. For instance, businesses can alert clients or storage facilities in advance of a truck's arrival, allowing for effective unloading and loading procedures.
    \item{Real-Time Inventory Tracking:}  GPTs enable businesses to manage their inventory levels while on the road with a cloud-based platform that makes it simple to access inventory data from anywhere in the world. Better inventory management and decision-making are made possible by this real-time accessibility. This ensures that the appropriate quantity of stock is accessible when needed to fulfil consumer requests, while minimizing carrying costs and preventing lost sales as a result of stockouts. GPTs can streamline inventory management procedures by eliminating the need for human data entry into spreadsheets, saving time and cutting overhead costs \cite{dwivedi2023so}. With the advent of 5G technology, the cost of connected devices has dramatically decreased, making it more practical and affordable for businesses to set up and operate connected inventory monitoring systems. This may make real-time inventory tracking solutions more widely adopted, thereby increasing the effectiveness and precision of inventory management \cite{zheng2021applications}.
    \item {Streamlining Delivery Operations:} GPTs are able to estimate traffic trends and improve routes for both drivers and passengers using real-time data \cite{haluza2023artificial}. These models can produce effective routes that reduce travel times and enhance overall delivery performance by taking into account aspects like traffic congestion, road conditions, and delivery schedules. Route optimization not only reduces travel time but also benefits the environment. In order to improve air quality and create a more sustainable delivery process, it is possible to cut down on idle times and trip distances. Businesses may streamline operations, improve the overall customer experience, and contribute to a more sustainable and environmentally friendly approach to logistics by automating procedures, optimizing routes, and utilizing real-time data \cite{rathore2023digital}.

    \item {Tourism:} GPTs have the potential to significantly improve a number of tourism-related aspects. GPTs can offer customized solutions that suit the individual's preferences by understanding their needs and interests, resulting in a more pleasurable travel experience. GPTs are excellent at understanding and creating text that is human-like \cite{lund2023chatting}. This functionality can be used in the travel and tourism sector to enable chatbots or virtual travel assistants to communicate with users in natural language \cite{bulchand2022impact}. Trip planning and information retrieval are made more simple and user-friendly by the ability of travelers to ask questions, look for advice, and obtain full details about destinations, modes of transportation, customs, and more. GPTs are capable of producing in-depth and interesting descriptions of tourist sites, attractions, lodging, restaurants etc. GPTs can provide time-efficient routes that guarantee a complete travel experience \cite{gillani2023unpacking}. Including advice on local legislation, emergency contacts, medical facilities, and potential risks, GPTs can offer helpful information and direction regarding travel safety. 

\end{itemize}
\subsubsection{Challenges}
Privacy issues may occur when using sensitive data in travel GPTs. It is essential to manage user data sensibly and putting up strong security measures to safeguard private data. The quality of the model's outputs is directly influenced by the correctness and completeness of the data utilized during the training phase. Ethical considerations should be taken into account when creating AI-powered applications employing GPTs. It's crucial to check that the models are truthful, unbiased, and free from harmful presumptions or discriminatory procedures \cite{mehrabi2021survey}. Although the models contain advanced features, they are difficult to tailor for specific use cases, need a lot of data to train, and have built-in limitations.

\subsubsection{Summary}
Emerging GPTs have the potential to enhance productivity, communication, and the calibre of goods and services, which will benefit many aspects of people's life. GPTs can offer real-time updates, effective route optimization, and customized recommendations in the travel and transportation industries, enhancing the overall travel experience and increasing operational effectiveness. Adopting them, however, comes with some difficulties. As specific roles are replaced by automation, GPTs may result in job displacement \cite{tschang2021artificial}. Additionally, the computational and memory requirements for GPTs make their deployment on compact or low-power devices difficult. GPTs may not be accessible to growing businesses due to the high costs associated with obtaining and using them. Despite these obstacles, attempts are being done to overcome them and improve the usability and value of GPTs for a larger range of users.

\subsection{E-Commerce\hl{}}
\subsubsection{Introduction}
Electronic commerce, commonly referred to as e-commerce, is a way for conducting economic transactions and create relationships between groups of people and entities using digital information processes and electronic communications \cite{jain2021overview}. Globally, this type of trade has experienced substantial growth, particularly in the retail sector. The preference for internet shopping, especially among younger millennials, is a prominent trend in consumer behaviour. Mobile devices have consequently taken over as the main method for carrying out internet transactions \cite{feijoo2020online}. Therefore, it is crucial for e-commerce companies to give the customer experience in their mobile applications top priority. The provision of brief text summaries for titles and reviews is an essential component of this. These summaries are essential for optimizing search results, helping consumers identify appropriate items, and ultimately raising customer happiness in the online purchasing space \cite{zhang2021dsgpt}.
\subsubsection{Impact of GPT in E-Commerce Realms}
The e-commerce sector could significantly advance with the introduction of GPTs. GPTs can be accessed by users or customers and are intended to answer commonly asked questions and give in-depth details about many elements of the e-commerce process, such as products, delivery, refunds, and more \cite{el2023sentiment}. One of the main benefits of GPTs is their capacity for quick responses, which decreases the amount of time customers must wait to hear back from businesses \cite{radford2019language}. By taking care of an important number of client inquiries, this function not only increases customer happiness but also lessens the workload on support workers. Customers will ultimately have a better purchasing experience as a result of being able to quickly acquire the information they require and interact with GPTs \cite{george2023review}.
\begin{itemize}
    \item {Proofreading:} To improve the calibre and accuracy of written content in e-commerce, GPTs can be used for proofreading. Written content is essential for product descriptions, marketing materials, customer reviews, and other text-based components in the e-commerce sector \cite{salminen2022creating}. For the purpose of projecting professionalism, fostering trust, and delivering a satisfying user experience, this text must be devoid of errors, well-written, and grammatically correct. E-commerce companies can automate the process of identifying and correcting these problems by using GPTs for proofreading, which saves time and effort as comparison to manual proofreading \cite{eloundou2023gpts}. This can be especially helpful in situations when users are writing product reviews or interacting with customer service. An improved user experience is facilitated by the early detection and rectification of errors, which also helps to avoid potential misunderstandings or miscommunications.
    \item {Order Processing:} GPTs are useful in many areas of order management and customer service because they can comprehend and produce text that looks like human speech.  GPTs can help with handling consumer questions about orders \cite{maddigan2023chat2vis}.  GPT is capable of interpreting the queries, providing important details like order status, tracking information, and expected delivery time, as well as suggesting corrections for frequent problems \cite{jain2022jigsaw}. By delivering real-time information, GPTs can assist customers in tracking their orders. Customers can customize their purchase with the help of GPTs. GPTs can help in the identification of possibly fraudulent orders by examining past transaction data, consumer behaviour patterns etc \cite{haleem2022era}. Based on a customer's past purchases, browsing habits, and preferences, GPTs can offer tailored product recommendations. When a consumer puts a purchase, the model can examine the information and produce recommendations for related or supplementary products that the customer might find interesting.
    \item {Generating titles for products:} Companies can use GPTs to produce interesting and educational material to improve the appeal of their product listings \cite{dwivedi2023so}. Based on a product's category, brand, and special characteristics, GPTs can come up with attractive titles for it. The model can produce imaginative and memorable names that aid in brand awareness and differentiation by receiving relevant information such as the characteristics of the product and the target market. GPTs are trained to produce in-depth and interesting product descriptions \cite{liu2023summary}. These summaries can offer a thorough summary that aids clients in selecting products wisely. GPTs are capable of coming up with clever and appealing captions for product images. GPTs can be adjusted to better reflect the tone and aesthetic of a certain brand \cite{brand2023envisioning}. As a result, the brand identity is consistent and unified throughout all product listings.
    \item {Strategy Planning:} GPTs have the ability to come up with original and distinctive concepts for marketing campaigns \cite{dehouche2021plagiarism}. The model can provide recommendations for different campaign aspects, such as slogans, taglines, themes, contests, social media strategies, and more by taking into account relevant information about the product, target audience, marketing objectives, and desired outcomes. GPTs can help with email writing that encourages readers to become partners, investors, or customers \cite{lund2023chatgpt}. To increase the likelihood of a favourable response or interaction, these emails can be customized to address the needs and potential benefits for the receivers. To improve their comprehension and production of appropriate material, GPTs can be trained on domain-specific knowledge bases, such as e-commerce \cite{el2023sentiment}. The models can offer more precise and situation-specific recommendations for advertising strategies, product positioning, and target audience interaction because of this specialized training. 
    \item {Data analysis:} There are numerous ways to use GPTs for data analysis in e-commerce. E-commerce data preparation can be aided by GPTs \cite{taecharungroj2023can}. Data normalization, cleansing, and formatting are a few of the duties involved. GPTs can produce summaries, identify significant topics, and extract appropriate data by studying textual descriptions, reviews, and consumer feedback. This helps you know the data more thoroughly, identify trends, and find insightful information. Customer reviews and social media comments from e-commerce can be analyzed for sentiment using GPTs. The sentiment expressed in text can be evaluated using GPTs and categorized as either positive, negative, or neutral \cite{el2023sentiment}. Understanding client views, recognizing product strengths and deficiencies, and making data-driven decisions all benefit from this analysis, which also helps to increase customer happiness. Segmenting consumers based on preferences, behaviours, or past purchases can be aided by GPT models \cite{javaid2023chatgpt}. For the purpose of detecting fraud in e-commerce transactions, GPTs can be used. GPTs can support the identification of potentially fraudulent actions by examining past transaction data, user behaviour patterns, and recognized fraudulent tendencies \cite{ray2023chatgpt}. 
\end{itemize}
\subsubsection{Challenges}
While GPTs have a lot of potential for numerous e-commerce applications, they also have several drawbacks. In order to produce responses, GPTs mostly rely on the context given in the input text. They could, however, find it difficult to fully understand the broader context or details that are unique to e-commerce. GPTs provide replies using training data and prior knowledge. They are unable to access real-time data or carry out real-time calculations \cite{wilcox2020predictive}. They might not be appropriate for giving current information, such as pricing, product availability, or dynamic promotional offers. GPTs gain their knowledge from a wealth of training data, which includes text taken from the internet, which may be biased, stereotyped, or otherwise offensive \cite{sridhar2022explaining}. The models may unintentionally provide biased or unsuitable replies if they are not rigorously managed and monitored, which could be harmful to the customer experience and brand reputation. The use of ethical principles and the training data must both be given careful thought.
\subsubsection{Summary}
The conversational interface offered by GPTs customizes the purchasing process and makes interactions with clients more interesting and appropriate to their individual requirements. GPTs can also be utilized to get insightful customer feedback. Businesses can learn about customers' preferences, issues, and opinions regarding their products and services by conversing with them. In order to better serve their target audience, organizations can use this information to discover areas for improvement, increase customer happiness, and make data-driven decisions. 
It's essential to recognize that GPTs might occasionally make mistakes or give poor answers, particularly when dealing with complicated or ambiguous queries. This highlights the necessity of continual model training, thorough testing, and modification to guarantee that they consistently meet consumer needs. To confirm the efficacy and dependability of using GPTs specifically in the e-commerce area, more research and testing are required. When implementing GPTs, it's critical for businesses to take into account the particulars of their own e-commerce businesses, their target market, and the type of the client enquiries. Regular monitoring and feedback analysis, along with a systematic and iterative approach, can help make sure that the outcomes of using GPT models are in line with the objectives of e-commerce enterprises.

\subsection{Entertainment\hl{}}
\subsubsection{Introduction}
In the ancient days, the Entertainment meant about playing games with neighbors covers all outdoor games, indoor games and chatting with neighbours through telephone. As digitization has bought greater advancements in computation and communication, in turn access to internet is also much easier. This has changed the way people are entertained. as people are connected and fully engaged in completing the target for the day. And there was a radical shift from traditional employment to employment in the Industrial Revolution age. Stress and pressure are common factors hindering people of different age groups. The different forms of entertainment serve as stress busters. Entertainment and mental health are interrelated; the former transfers happiness, bringing harmony and peace to mental health. Some common forms of entertainment include playing games, watching TV series or movies, or funny videos, shopping, debugging, coding, browsing the internet, listening to music, dancing, chatting, painting, crafting, reading books, cooking, and many more, which can lessen the stress carried\cite{bryant2013psychology}. Entertaining and getting entertained is the biggest motivation and medicine for all mental illnesses. Entertainment helps to improve the motor skills of humans, thereby inducing a positive cognitive effect towards the work.
\subsubsection{Impact of GPT on the Entertainment Industry}
GPT is a potential game-changer in the entertainment field, delivering endless entertainment. Since its evolution, GPT models have been adopted as an entertainer crosschecking their ability to produce content on funny and illogical questions. GPTs entertain people in many ways, and of course, using GPT itself an entertainment as it reduces the burden of overthinking by providing immediate feedback to queries in seconds\cite{ thorp2023chatgpt}. The results are amazing and have been utilized for many purposes today. When the GPT model was probed to complete a scene from the movie “Frozen,” it responded with an entertaining writeup\cite{shahriar2023let, haque2022think}. Some of the impacts of GPTs on Entertainment applications are given below:

\begin{itemize}

\item {Solitude with GPT:} As the GPT itself is an entertainer, one can feel better alone with the GPT, which helps to come out of loneliness by exploring its savors\cite{ai}. GPTs assist in providing soothing poems, mental healing quotes, and funny riddles. People with loneliness may feel anxiety, especially with older ones at home. In this case, GPT-4 helps people with its Voice Technology feature, enabling users to input their audio\cite{ai}. In turn, the GPT model responds to user-specific speech output using NLP algorithms embedded with it. The elderly can feel safe and attentive at home. GPT-4 is multilingual and can understand various dialects and accents for personalized user experience.

\item {Enhanced Customer Interaction:} The advent of ChatGPT and Bard has improved customer interactions on content such as movies, Over-the-Top (OTT) platforms like Netflix, Hulu, Disney+ Hotstar, and prime video, sound recordings, song lyrics, pictorial works, comics, jokes, memes, viral videos, and other entertaining factors. Further, GPTs provide human-like recommendations on user-specific fun activities based on user interactions for an immersive experience. This has dramatically improved the interactions in the engagement industries. User engagement can be further improved by providing dynamic and more realistic responses to user queries, such as creating virtual actors for interacting with real actors\cite{opchatsgpt_2023}.

\item {Personalized Content Creation:} GPTs can help generate user-specific personalized content by analyzing the user preferences and generating content like predicting future scenarios tailored to the user’s interest. GPTs can be used for creating personalized, engaging, and high-quality content for online business advertising, ideas for content generation, marketing messages for attracting customers, descriptions for selling products, and captions for social media\cite{linkedinPowerChat}. In addition, it can be used for optimizing the contents for search engines, i.e., GPTs will provide relevant terms for search, thereby avoiding traffic to the web sources.

\item {For the Film and TV industry:} GPT-powered virtual assistants assist users in booking tickets and generating content and personalized recommendations using AI models. The evolution of GPT-4 with advanced NLP and DL algorithms helps the scriptwriter to generate AI-driven content without the human author named virtual storytelling\cite{ts2ChatGPT4}. GPTs create interactive stories, dialogues, and characters, recommending suitable characters. Furthermore, GPTs can be used to create content for video games, voice-enabled applications, AR applications, and other VR experiences in virtual worlds\cite{ bd_2023}.

\item {For Social media influencers:} GPTs can generate personalized marketing ads for each customer based on their previous interactions and provides relevant suggestions for customer viewing experiences. Youtubers and other social media content creators will potentially benefit from generating channel content based on demand and realistic societal activities.

\item {Realistic Gaming Interactions:} GPT helps to generate the players, gaming narratives, dialogues, user interface, and user-specific gaming recommendations and new game creation. Powerful HCIs can render a better user experience for game developers and players. Assistance to the game developers in debugging and enhancing the code developed. GPT uses various NLP and AI algorithms trained with massive data to predict the next phrase/movements and provide human-like experiences in 3D gaming environments. ChatGPT has been integrated with AR and VR to provide an immersive gaming experience.

\end{itemize}

\subsubsection{Challenges}
Latency is the major issue connected with rendering the voice-based response to the voice input. As well, plausible misinterpretations may mislead the responses, and interruptions to the relayed output are difficult. Enabling technologies like EC and 5G can help overcome this issue. Also, GPTs must be capable of storing the facts with audio conversations to relay them while conversing the other day. Furthermore, the AI system must be built in such a way that it can continuously learn (lifelong machine learning) and enhance over time. The major ethical concern with virtual storytelling is the bias exposed in the training data and the obscurity of reproduced content on the generated stories. Another issue with the generated content is plagiarism (i.e., producing content similar to the content in the published articles or books), raising disputes with intellectual property rights. In addition to this, the source of the content generated remains unexplored. The language barriers in using GPT must be lessened to improve user experience and utilize the features of GPT \cite{liu2023summary}. The implication of the user to provide inputs in a certain format to GPTs can be further improved by providing different options in addition to voice-based inputs GPT4, like braille screen input for visually disabled people. The user authentication can also be further enhanced to safeguard user-specific content generation and avoid repeated content generation for users with similar requests. One of the primary concerns with the GPTs adoption is job loss. Content creators, bloggers,  and poets may lose their jobs.
\subsubsection{Summary}
The entertainment industry is the one which will be in demand always, as it is a lifeline for many individuals leading a stressful work environment or personal life. Despite the stress, entertainment has become part of routine life due to its immersive nature, creating harmony in the mind and the environment. GPTs have made a major contribution to enhancing the entertainment industry, but the job security of many professionals in this field remains unanswered. GPTs must be trained on unbiased data and ensure transparency in source content generation to provide a secure, robust, and efficient contribution to the entertainment industry. To attract all types of users, the multilingual capability and content rendering of GPTs can be further enhanced. The issues constrained by providing user inputs to GPTs can be alleviated to all extent.  Furthermore, safer user content generation without plagiarism and relating facts with previous conversations can be guaranteed by abiding by the storage requirements to deal with a more personalized user experience.
\subsection{Lifestyle\hl{}}
\subsubsection{Introduction}
Lifestyle, the way of our living, is one of the prominent areas most people in today’s digital era of AI,  are bound to and look for constant improvement. The “modus vivendi” is a Latin expression that semantically means a way of living and should be understood in terms of values and attitudes. These two terms manifest self, influenced by family, society, and global media. Directly or indirectly, these influence an individual’s lifestyle. Adopted from a sociological perspective, an individual expresses oneself through different practices, viz., eating ways, drinking behaviours, mode of travel, travelling places, costume designs, body-shaping cloth to wear, media preferences, education choices, job preferences, entertainment modes, managing leisure time, means of communication and so\cite{veal1993concept}. In all these practices, individuals would like to explore and learn about what, where, how, and when factors for reading sustainable development\cite{ jensen2007defining}. The concept of lifestyle is all about “how one wants to live one’s life.”  Consumerism is the act of purchasing artifacts for societal status and is one of the thriving lifestyle factors. Certain standard indicators like job, wealth, and physical and mental health determine the quality of one’s life. Also, the choice of a healthy lifestyle moderately determines the health of an individual\cite{ contoyannis2004socio}. Furthermore, few people believe that lifestyle reflects their socioeconomic status. Many epidemiologic studies state that better lifestyles have dramatically reduced the risk of various chronic diseases and are the primary cause for their prevention\cite{reeves2005healthy}. The lifestyle has been defined on different societal levels from individual, positional, and national to global\cite{ jensen2007defining}.  At the global level, lifestyle is adopted by general world-class influencers. In contrast, at the national level, the influencing factors will be the government and different cultural patterns across the country. The positional level concerns influence from different status groups, age categories, gender groups, and social classes. And the individual level is influenced by a closely moving group of individuals concerned about self-identity. The major source of information about these influencers is the Internet through social media networks and personal development advertisements. 
\subsubsection{Impact of GPT in Lifestyle}
The most remarkable application of AI, the GPT, paves the way for the betterment of mankind in offering human-like intelligent conversation on all whereabouts. People will always prefer to interact with other peers to learn their attributes and tweak them for societal status. Various GPTs have flourished for different lifestyle indicators, and they provide human-like assistance to all queries on fine-tuning the lifestyle by harnessing the power of AI\cite{yeo2023assessing}. The advanced reasoning capability of GPT-4 serves the purpose better\cite{ai}.
\begin{itemize}
\item {Diet Planner:} Free GPT applications for maintaining a balanced diet, helping the individual with a weight loss diet plan, followed by a brief list of meal plans, required shopping lists, physical activity plans targeting particular body parts\cite{biswas2023role}, motivational messages, and personalized sleeping patterns.  These apps act more like personalized training assistance and help to track progress with visualization charts or graphs. Fitness level, available free hours, medications taken, and available exercise equipment will be given as input to GPT.
\item {Travel Guide and Trip Advisor:} Harnessing AI models, GPT provides an individual’s travel plan itinerary based on information like the place(multiple cities), budget, and the number of days. These GPTs provide local recommendations on restaurants, hotels, and other attractions. RoamAround, Roamr, and VacayChatbot are some of the travel planning GPTs\cite{makeuseofFreeTravel}.
\item {Personalized Stylist and Beauty Advisor:} GPTs can act as personalized stylists for an individual by generating occasion-specific clothing and costume preferences.  GPTs can assist in organizing wardrobes based on seasonal outfits and provide recommendations on e-commerce fashion stores for purchasing favourite brands. GPTs can provide tailoring design options, fabric choices, and design materials. Furthermore, GPTs can provide updates on a stock based on the preferred searches and provide insight into fabric types suitable for weather conditions that suit personal style.
\item {Personalized CookBook:} GPTs can serve as cooking assistants by recommending new curated recipes suiting the family dietary plan, ingredients available, time, individual’s cooking skills, and new flavoured ingredients.  ChefGPT, PantryChef, and MacrosChef are some GPTs that generate unique and delightful recipes\cite{chefgpt}. Consequently, GPTs can assist in shopping list recommendations and the nutritional value of the recipe generated.
\item {Hobby Curator :} GPT assists an individual in identifying one’s enjoyable leisure time activity by learning new skills\cite{lc}. Having a list of interests and ideas ready, the GPT helps narrow down various options, instructional videos to proceed, chatting and sharing with online communities, and researching the cited hobby to explore more fun. Budget will also be an important factor in this perspective, as learning new hobbies may require joining paid classes or courses. GPT provides step-by-step instructions and guidelines to learn a new skill faster.
\item {Dream Maker:} GPTs with multimodal learning helps to search for a job based on one’s qualifications and experience. In turn, it assists in preparing the job-specific resume, cover letters, training for the interviews (coding and technical queries), and grooming sessions and can redirect to the training place where knowledge can be acquired\cite{indeed}. The futurist GPT models can assist in phase-by-phase questionnaires in the interview process
\end{itemize}
\subsubsection{Challenges}
The recent version of the GPT uses both reinforcement and supervised learning models so it can learn based on the interaction with the user and can use existing data to derive personalized decisions. In the context of lifestyle, GPTs offer the most promising solution for almost all lifestyle influencers, but the still challenging part is the trustworthiness of the data and copyright issues. Also, relying more on GPT as it solves all our problems may insipid human intelligence in upcoming generations. Though the GPT provides weight advice, it can never be a substitute for the medical practitioner, as some information can be misleading. Travel planning GPTs sometimes require users to update information in a specific format and may have outdated databases. GPTs cannot access specific job openings' websites but can still provide insights into acquiring them. At times, it can produce nonsensical information \cite{floridi2020gpt}. Therefore, before adopting the GPT recommendations fully, further instigation is recommended. Furthermore, developing a large multimodal learning model abiding huge and dynamic datasets will be costly. 
\subsubsection{Summary}
GPT is a personalized assistant for improving an individual’s lifestyle from various prospective influencers. Generating personalized recommendations alleviates an individual’s fear of survival in the digitized society. Individuals will be personally trained to adapt to different cultural and technological shifts in the sustainable development of themselves and the economy as a whole. On the other hand, more stringent recommendations may incur huge budget overruns and sometimes provoke the individual to misinterpret, leading to dreadful consequences. GPTs provide both positive and negative recommendations based on the input fed. So, for the effective adoption of a GPT for lifestyle practices, adverse training and testing on extreme behaviours must be carried out. GPTs must be trained in the realistic and dynamic perception of individuals in real life.

\subsection{Gaming}
\subsubsection{Introduction}
Before the advent of technology and the gaming industry, entertainment was primarily centred around activities such as reading, listening to music, watching plays and movies, participating in sports and physical games, and socializing with friends and family. People also engaged in traditional board games and card games, which were often played in groups and provided a fun and social way to pass the time. After technology stepped into the gaming industry, the way games are created, and the experience it has given users have transformed tremendously. Technology has enabled developers to create more immersive and engaging experiences for players. It has contributed in various ways, like improving graphics, performance, online play, and mobile gaming. Improved GPUs and other technologies allow for more detailed and realistic graphics, making games more visually stunning. Faster processors and higher amounts of RAM allow for smoother gameplay and faster loading times, reducing lag and improving overall performance. Technologies like AI, AR, and VR have created a new dimension of game development and experience. Players can now immerse themselves in gaming worlds in a way that was not possible before. With the help of advanced AI techniques, game developers can create more sophisticated and challenging opponents for players, as well as NPCs with more realistic behaviours. Technology has greatly expanded the possibilities of gaming and enabled developers to create more immersive, visually stunning, and engaging experiences for players.
\subsubsection{Impact of GPT in Gaming }

GPTs have the ability to contribute to all sectors, including the gaming sector. GPT are not specifically designed for creating and playing games, but they have the potential to improve the gaming experience by improving enhanced dialogue and story telling, creating dynamic and personalized gaming worlds, generating more realistic and engaging characters \cite{toshniwal2021learning}, game content creation, chatbot development. 

\begin{itemize}
    \item {Chatbot development:} GPTs have been used in gaming through the development of chatbots that use NLP to communicate with players \cite{freiknecht2020procedural}. Because it allows the chatbot to understand and respond to a wide range of user inputs and queries related to the game. GPTs have been pre-trained on a large corpus of text data, which makes them adept at NLP. It can understand and respond to user queries in a way that feels natural and intuitive. It can also understand the context of a user's query, which means they can provide relevant and useful responses even when the user's query is ambiguous or incomplete. It can also generate game-related content, such as descriptions of game characters or settings, that can help to enrich the user's gaming experience. Furthermore, it can also personalize the user's experience by learning from their previous interactions with the chatbot and tailoring its responses accordingly.
  
    \item {Game content creation:} GPTs are used in game design. They are used to create game content such as levels, items, and quests. If the game designer is working on a new role-playing game, GPT can be used in creating characters to be used in the games. To generate new character classes in the games, the developer has to give inputs that contain information about the game environment, game settings, player abilities, and game play mechanics. GPTs has the ability to analyze the text and expectations given by the developer, and it can generate a list of potential character classes based on the expectations given as text. The designer then refines the ideas and chooses a more suitable character to develop further with unique abilities and game mechanics. The authors in \cite{9980408} have used GPT2 and GPT3 to procedurally generate role-playing game with video game descriptions. The resultant quest  was evaluated by 349 online RPG players. The results concluded that one of the five quest descriptions was accepted for game development.
    \item {Analyze player’s ability and skill:} GPTs can detect and analyze players' abilities and skill levels and tailor the game accordingly. This analysis helps in making dynamic modifications to the game environment based on the player's abilities and skill levels. This feature helps achieve dynamic difficulty balancing. GPTs can also assist in identifying the player's intent. Thus, when players ascend to higher levels, it can assist in making the games more challenging based on the player's abilities and skill levels in the previous levels
    \item{NPCs:} NPC stands for "Non-Player Character." In AI games, NPCs refer to characters or entities in a game that are not controlled by a player. NPCs can take on a variety of roles within a game, such as enemies to fight, quest givers, merchants, or friendly characters that provide helpful information. They are often controlled by AI algorithms that determine their behaviour and actions within the game world. GPTs are not specifically designed for creating NPCs, but they can be used to generate dialogue and other character interactions that can be incorporated into NPCs. Additionally, It can be used to generate character backstories and personalities, which can inform the development of NPCs. The authors in \cite{10.1145/3472538.3472595} have trained and used GPT-2 for text generation of video games. They have trained GPT-2 on a large corpus of video game quests and used a GPT model to generate the dialogue for quest-giver NPCs in role-playing games. The output has shown that GPT can learn the structure and linguistic style of the games, and the quality of the content it has generated is high, making it a good alternative to writing new RPG quests by hand.
    
\end{itemize}
\subsubsection{Challenges}
 GPTs are computationally expensive and require high computing resources to do their purpose. This means that implementing them in a game would require powerful hardware and this could have an impact on the performance of the games.
Lack of training data: GPTs require large amounts of high-quality training data to be effective. In the gaming industry, this could be difficult to obtain, as gaming data are likely to be fragmented and less structured than the kind of data used to train GPT models \cite{radford2019language}. In addition, GPTs can perform content creation based on patterns they have learned from their training data, which means that they can be unpredictable. The content generated by GPT may be nonsensical or inappropriate content to the game. In the context of gaming, this lack of control could lead to undesirable or even offensive game content. GPTs can generate text based on user input, they can't interact with the game environment in the same way a human player can. This limits their usefulness in gaming and may make them less effective than other AI technologies.

\subsubsection{Summary}
GPTs can transform the gaming industry by contributing to improved game dialogue creation, enhanced non-player characters, personalized gameplay, procedural content generation, chatbot generation, and analyzing players' abilities. However, it also has potential challenges that are to be addressed, such as the need for high computing resources, a lack of control over content creation, and restricted interaction with the game environment. In addition, the most important challenge in adopting a GPT model in gaming is a lack of training data. If the challenges are addressed and the gaming industry evolves with properly structured data to train a GPT model, then GPTs can revolutionize the field of gaming.
\subsection{Marketing}
\subsubsection{Introduction}

Traditional marketing primarily relied on traditional media channels, such as television, radio, newspapers, and magazines, to reach consumers. Companies used to develop marketing campaigns based on demographic data, and mass media channels were used to broadcast these campaigns to a broad audience. However, the advancements in technology have brought about significant changes in the marketing industry, and companies are increasingly integrating new marketing strategies evolved through technologies to reach and engage with customers. One of the significant transformations has been the rise of digital marketing channels such as social media, search engines, Email, and mobile applications that allow companies to target specific populations with precision and provide real-time feedback on campaign performance, allowing for more effective and efficient marketing. Technology has also given rise to marketing automation tools such as customer relationship management systems, chat-bots, and personalized email marketing, which have made marketing more efficient and effective. Another significant transformation has been the use of big data and analytics to better understand customer behaviour and preferences. This has allowed companies to create more personalized and targeted campaigns based on specific customer needs and preferences.

\subsubsection{Impact of GPT in Marketing }
The marketing industry has evolved with various AI-powered techniques. This revolution started in marketing by providing businesses with powerful tools for generating insights, automating processes, and improving customer experiences. GPTs are also being used in marketing to generate engaging and personalized content. Some of the applications of GPT in marketing include content creation, customer service, and personalized advertising.

\begin{itemize}
    \item {Content creation:} GPTs can contribute to marketing in various ways, such as by improving speed and efficiency in content creation, ensuring consistency and quality of content, generating personalized content, creating multilingual content, and repurposing existing content. It can be trained on a company's existing marketing materials and customer data, allowing it to create new content, such as blog posts, social media updates, and product descriptions, in a fast and efficient manner. Despite its speed, it maintain high standards for quality and consistency. Moreover, GPTs \cite{floridi2020gpt} can generate personalized content based on customer data, such as search history and past purchases. This helps create content that is relevant to the users' desires, leading to better engagement and conversion rates. GPTs can also generate content in various languages, allowing marketers to expand their reach across regions. Copy.ai \cite{roetzer2022marketing} has used GPT-3 to generate human-like text that is optimized for marketing purposes such as website copy, social media posts, advertisement copy, and email campaigns. This means that marketer personnel no longer focus on content creation. Instead, they can spend productive time improving the other aspects of marketing. 
    \item {Customer service:} GPTs can be trained on customer service conversations and chat logs to generate more natural responses, like humans. This can help business personnel provide better customer service 24/7 and save time and resources. It can be trained to generate automated responses for frequently asked questions, providing faster responses to customers and ensuring consistency in the quality of replies. GPTs can also analyze customers' emotions and sentiments, enabling businesses to proactively address negative feedback. This is particularly helpful in maintaining customers' trust. The authors in \cite{thiergart2021understanding} have used GPT-3 model for automated drafting of responses for incoming mails. They used it to understand the mail, and then software engineering and business studies were used to understand the challenges encountered and finally, the response generated after a thorough understanding of the context of the mail. The authors have concluded that applying GPT-3 to rationalize email communication is feasible both technically and economically.
    \item {Personalized advertising:} GPTs can generate personalized content such as product descriptions, blog posts, and social media captions tailored to individual customers' preferences and interests. This can help businesses create content that resonates with their target audience, leading to higher engagement and conversion rates. By analyzing customer data, GPTs can segment customers according to their behaviour, interests, and preferences. As a result, businesses can tailor their marketing campaigns to each segment and provide personalized messaging and offers that are more likely to connect with each customer group. The authors in \cite{10.1145/3511808.3557129} have proposed a generative model to identify the name of the product from the product text and use this information filter to improve the product recommendation based on the product retrieval model. This method has been implemented in the dynamic product advertising system of Yahoo. It is observed that the recommendation system has recommended the product based on the user's interest, and it was evaluated using an A/B test to serve similar products in an ad carousel, which can help the system to explore more products efficiently.
    \item{Forecast analysis:} Using customer data analysis, GPTs can forecast future behaviour and buying patterns. This allows businesses to customize their marketing campaigns to each customer's desires based on their purchase patterns, increasing the likelihood of conversion or purchase. The authors in \cite{neves2022chat} have used chatGPT to perform predictive modelling based on past data. They have used the GPT model to predict the future based on the customer’s behaviour and buying pattern. This primarily helps the system to recommend the products to the customers as per their desires. 
\end{itemize}
\subsubsection{Challenges}
GPTs are designed to generate content that imitates human writing, but the content generated may not align with the brand's image or message. This lack of control can be a potential challenge for marketers. Another challenge that applies to all learning technologies is that data bias is possible in GPTs \cite{lund2023chatgpt}. Based on the large dataset of text used for training, if the data is biased, it will affect the generated content, which may also exhibit the same biases. GPT is complex and difficult to interpret, making it challenging to explain how the model arrived at its conclusions. This lack of transparency can lead to a lack of trust in adopting GPTs, and marketing teams may struggle to make improvements in their strategies. As like every AI technology, there are ethical concerns associated with GPT models. For instance, the use of GPT in marketing could raise concerns about the use of personal data and privacy, particularly if the model is used to generate targeted advertising or personalized content. To avoid any negative consequences, companies must ensure they use these models ethically and transparently.
\subsubsection{Summary}

Using GPTs in marketing can provide various benefits, such as better content creation, personalized messaging, increased efficiency, competitive advantage, and enhanced customer experience. However, this strategy also involves potential challenges, such as limited control, data bias, lack of transparency, and ethical considerations. Therefore, companies must consider the advantages and drawbacks of GPT adoption in marketing, and implement these models ethically and transparently to avoid negative outcomes. Successful integration of GPTs in marketing requires proper planning, a skilled workforce, and continuous monitoring to ensure the desired results and mitigate any potential risks.
\subsection{Finance}

\subsubsection{Introduction}
The finance industry, also known as the financial sector, is a broad term that encompasses a wide range of institutions and businesses that provide financial services to individuals, businesses, and governments. The finance industry plays a critical role in the global economy, facilitating the flow of funds between savers and investors, managing risk, and providing financial services and products to support economic growth. The finance industry has been the leader in technology adoption in recent years, with a focus on improving efficiency, reducing costs, and delivering better customer experiences. The adoption of technologies like big data and analytics, mobile and digital payments, blockchain and distributed ledger technology, AI and ML, and cloud computing make the sector more flexible, scalable, trustworthy, transparent, secured, and easier to access.
\subsubsection{Impact of GPT in Finance}
GPT has greatly influenced finance by automating customer support using chatbots and virtual assistants, enhancing fraud detection, offering investment insights and recommendations based on financial data and news, assisting with risk assessment for investments and loans, impacting algorithmic trading strategies, simplifying compliance with regulations by analyzing legal documents, improving credit scoring and loan processes, and emphasizing the importance of handling sensitive financial data securely and transparently.

\begin{itemize}
    \item {Sentiment analysis:} Sentiment analysis is a technique used in the finance industry to evaluate the sentiment of investors \cite{gong2019sentiment}  and the general public towards specific companies, industries, or markets by analyzing news articles, social media posts, and other text-based sources of information. GPT has the potential to improve sentiment analysis in finance by providing more accurate and detailed analyses of financial data. With sentiment analysis, the industry can predict stock prices by assessing the sentiment of news articles, social media posts, and other sources of information about a particular company or industry to make informed investment decisions. By utilizing sentiment analysis, GPTs can aid financial institutions in identifying potential risks and taking appropriate action to mitigate them. The authors in \cite{sweidan2021sentence} have investigated how incorporating a lexicalized ontology can enhance the performance of aspect-based sentiment analysis by extracting indirect relationships in user social data. The investigation results show that the analysis has given 98\% accuracy.
    \item {Financial forecasting:} GPTs can be trained on past financial market data to predict future trends in the stock market, exchange rates, and other financial metrics. This can help investors and financial organizations make more accurate predictions and reduce their risk exposure. With the ability to analyze and process the natural language, GPTs can be used to analyze and interpret financial data, news, and other related information. Financial analysts and researchers can use the ability to analyze natural language to extract insights from unstructured data like news articles, social media content, and other information that is relevant to forecasting. This can help improve the accuracy of financial forecasting models by providing a more comprehensive view of market trends and sentiments. This analysis may help improve the accuracy of prediction. Financial analysts can use the model to identify the relationship between the financial parameters that could change the market conditions in advance. This prediction may be helpful for investors as they make investment decisions.
    \item {Trading strategies:} GPTs can also be used to analyze market trends and historical data to develop trading strategies. This can help traders make better decisions in terms of trading to increase their profitability. GPTs can be used to identify the potential risks in trading portfolios. By analyzing the large volume of information related to trading, GPT will get the potential to identify the risk parameters and provide insights into how to mitigate these risks. The authors in \cite{wei2018stock} have used a popular GPT for stock market trend prediction. The results show that the method used is simple but the efficiency and accuracy of the method are very effective. The prediction it has made is very close to the reality. 

\item{Risk prediction and management:}The adoption of GPT can enhance the process of risk prediction and management in several ways. It can improve data analysis by detecting patterns that may pose a risk. It can also help in enhancing fraud detection by analyzing transaction data and identifying fraudulent activity based on patterns. Additionally, GPT can be utilized to make better portfolio management decisions by analyzing historical industry data, company financial statements, and news articles, as well as social media feeds. This portfolio management process can provide valuable information about the investment risk of a given organization, enabling informed investment decisions and effective risk management. 
\end{itemize}
\subsubsection{Challenges}
GPTs have more challenges in the finance sector. Primarily, they demand significant computational resources to train and deploy, which can be expensive and time-consuming for financial organizations to implement. Another challenge is that, even though GPTs are capable of producing precise predictions, they can be challenging to interpret, which can present a problem for financial institutions seeking to comprehend the reasoning behind specific predictions \cite{yue2023democratizing}. This lack of interpretability can harm risk management objectives.Implementing GPT in finance sector can be vulnerable to adversarial attacks, which are designed to manipulate the model's output by injecting false data. This can be particularly problematic for financial institutions that rely on GPTs for risk management and investment decisions. It also require large amounts of training data to achieve high accuracy. However, in some cases, financial institutions may not have access to sufficient data to train the model effectively. GPTs can also be biased if the training data used to develop the model is biased. This can lead to inaccurate predictions and unintended consequences.

\subsubsection{Summary}
The use of GPTs in the finance industry has promising benefits such as improved risk management, enhanced fraud detection, better portfolio management decisions, and increased efficiency. However, it also has potential challenges that need to be addressed, such as high computational requirements, the complexity of implementation, limited interpretability, vulnerability to adversarial attacks, limited training data, and bias in training data. So, the use of GPTs in the finance industry presents significant benefits but also requires careful consideration of the challenges involved to ensure the effective and secure deployment of these models.

\subsection{Summary On Impact of GPT models in Applications}
The impact of GPTs in various applications and challenges was highlighted. GPT  with its varied usage has changed the way people perceive facts such as content creation, enhanced user interfaces, personalized learning, item tracking, self-awareness, market risk analysis, business forecasts and introspection. However, there are concerns about the potential negative impact of GPTs, such as the spread of fake news, bias in data and decision-making, not domain specific, ethical issues, data reliability, the complexity of implementation, multimodal and multilingual support, security and privacy concerns, vulnerable to data attacks, limited input data, explainability of results, large model size, high computational requirements and job loss. Despite these concerns, it is clear that GPTs will continue to be a powerful tool for industries seeking to leverage the power of NLP and generative AI. As the technology improves and new applications emerge, it will be interesting to see how GPTs continue to shape the future of industries around the world.

\section{projects}

\begin{table*}[h!]
\centering
\caption{Project Summary Table.}
\label{tab:project1}
\resizebox{\textwidth}{!}{%
\begin{tabular}{|p{2cm}|p{3cm}|p{3cm}|p{3cm}|p{4.5cm}|p{4.5cm}|p{4.5cm}|}
\hline
\rowcolor[HTML]{EFEFEF} 
Project &
  DeepScribe &
  Meena &
  Jukebox &
  Uber's plato research dialogue   system &
  Polyglot AI &
  SiriGPT \\ \hline
Application widely used for &
  Healthcare &
  Lifestyle &
  Entertainment &
  Transport &
  Education &
  Lifestyle \\ \hline
Purpose &
  Medical documentation and to improve doctor-patient association &
  Personalized product recommendation &
  Enables the original music    creation both artistically compelling and commercially viable in a   variety of styles and genres &
  Enhances user experience using Uber rides, helps drivers and riders in   scheduling rides, navigating routes, providing real-time updates on traffic   and weather conditions. &
  enables absolute communication irrespective of the language barrier   across  different regions and   cross-culturalism &
  Assist with voice-based assistants \\ \hline
GPT Adoption &
  Customized version of GPT's &
  Google's seq2seq transformer-based neural network architecture similar to   Open AI's GPT &
  GPT-2 extension called "Multi-Scale Transformers for Music   Modeling" (MST) model &
  GPT-2 &
  GPT-0, GPT-1,GPT-2,GPT-3 &
  GPT-3 \\ \hline
Dataset &
  Not Disclosed &
  Meena dataset over 40 billion words , 341 GB captured from public domains   like Reddit and social media platforms &
  1.2 million songs, 600,000 pieces of sheet   music, 45,000 MIDI files &
  Persona-Chat with 160,000 conversational dialogues, Cornell Movie-Dialogs   Corpus with 200,000 movie conversation, DailyDialog over 13,000 dialogues,   and &
  CONLL-2003, Sentiment140 dataset, Reuters Corpus, 20 Newsgroups dataset,   WMT (Workshop on Machine Translation) datasets and SQuAD (Stanford Question   Answering Dataset) &
  Information not publisized \\ \hline
Building Blocks &
  Recurrent Neural Network and Attention mechanism fueled by NLP techniques &
  Seq2Seq Transformer-based Architecture &
  Transformer-based Language Model and Autoregressive model &
  Language modeling, Dialogue modeling, Discrete latent variabe modeling   and response ranking &
  Language Identification, Named Entity Recognition (NER), Sentiment   Analysis, Text Classification, Machine Translation, Question Answering &
  Transformer-based neural network architecture \\ \hline
Evaluation Metrics &
  Bleu score, perplexity &
  Bleu score, perplexity &
  Frechet Audio Distance (FAD) and Pitch and Rhythm Similarity &
  Bleu score, Perplexity and Distinct n-gram &
  accuracy, precision, recall, , F1-score, Bleu score as well as   cross-entropy loss or perplexity &
  Perplexity, BLEU score, F1 score, ROUGE score, Human evaluation \\ \hline
Addressed Challenges &
  Reduced Transcription errors and enhanced patient care &
  Natural and Engaging conversations &
  Fresh orginical music content creation and drastically reducing the cost   and time by creating high-quality music contents, and also to preserve and   advance musical heritage. &
  customer service, user experience, and operational efficiency &
  Multilingualism and Sentiment Analysis are the key challenges in NLP and   Polyglot AI solved this problem by offering a tool for supporting morethan 40   languages and pre-trained sentiment analysis model &
  Language understanding and generation, Data scarcity, Contextual   understanding, Text summarization, Sentiment analysis, Named entity   recognition \\ \hline
Input data &
  Audio &
  Text &
  Audio &
  Text &
  Text &
  Audio \\ \hline
Owned By &
  DeepScribe &
  Google &
  OpenAI &
  Uber &
  Uizard Technologies &
  Apple \\ \hline
\end{tabular}%
}
\end{table*}

\begin{table*}[h!]
\centering
\caption{Project Summary Table (continued).}
\label{tab:project2}
\resizebox{\textwidth}{!}{%
\begin{tabular}{{|p{2cm}|p{3cm}|p{3cm}|p{3cm}|p{4.5cm}|p{4.5cm}|p{4.5cm}|}}
\hline
\rowcolor[HTML]{EFEFEF} 
Project &
  AI Dungeon &
  Copy.ai &
  Bond AI &
  Viable &
  AI Channels &
  Fireflies.ai \\ \hline
Application widely used for &
  Gaming &
  Business and marketing &
  Finance &
  Business Analytics &
  AI Industry &
  Business \\ \hline
Purpose &
  Interactive and engaging storytelling experience for players &
  help clients create written content more quickly and easily &
  To enhance the financial well-being of clients &
  provide businesses with intelligent insights to help them make better   decisions &
  provide a platform for developers, data scientists, and machine learning   practitioners to create, deploy, and manage their AI models &
  to simplify the meeting process and reduce the time and energy required   for note-taking and collaboration \\ \hline
GPT Adoption &
  GPT-3 &
  GPT-3 &
  GPT-3 &
  GPT-4 &
  GPT-3 &
  GPT-4 \\ \hline
Dataset &
  Common Crawl, OpenAI GPT-2, and various text datasets from Kaggle &
  books, articles, and websites &
  likely use of a combination of publicly available financial datasets,   proprietary data, and client data &
  Information not publisized &
  Users' own dataset &
  Possible datasets: the Common Voice dataset from Mozilla   having over 9,000 hours of speech data in multiple languages \\ \hline
Building Blocks &
  Machine Learning Models, Text Input Interface, Game Engine, Content   Database, Player Feedback System, Cloud Infrastructure &
  NLP, Language Models, Neural Networks &
  NLP, Personalization, Conversational User Interface, Data Analytics &
  Unsupervised learning, Contextual understanding, Sentiment analysis,   Topic modeling, Entity recognition &
  Pre-built models, Model training, Data preparation, Collaboration &
  Speech-to-Text Technology, NLP, Cloud Computing, Integration technologies \\ \hline
Evaluation Metrics &
  Response Coherence, Response Diversity, Player Satisfaction, Engagement,   Realism, Novelty &
  Perplexity, BLEU score, ROUGE score, F1 score &
  Intent recognition accuracy, entity extraction accuracy, and language   model perplexity &
  Perplexity, Accuracy, F1 score, Word similarity &
  Accuracy, Precision and Recall, F1 Score, Perplexity, User satisfaction &
  Speech Recognition Accuracy, NLP Performance, Integration Performance,   Task Completion Time, User Satisfaction \\ \hline
Addressed Challenges &
  Narrative Generation, Content Creation, Personalization, Replayability,   Accessibility, Creative Expression &
  Lack of writing skills, Inconsistency, Multilingual content creation &
  Personal financial management, Customer engagement, Fraud detection and   prevention &
  Understanding unstructured data, Contextual understanding, Visualization   and exploration of data, Customization and integration &
  Natural language understanding, Scalability, Personalization, Integration   with other systems, Maintenance and updates &
  Time-consuming manual note-taking, Difficulty in capturing important   details, Lack of visibility and accountability, Communication barriers \\ \hline
Input data &
  Text &
  Text &
  Audio and Text &
  Text &
  Text &
  Audio \\ \hline
Owned By &
  Latitude &
  Copy.ai &
  Bond.AI &
  Viable AI &
  MiroMind AG &
  Fireflies AI \\ \hline
\end{tabular}%
}
\end{table*}

This section presents the exciting projects developed using GPT model technologies for different applications mentioned in the above sections. Table. \ref{tab:project1}, Table. \ref{tab:project2} shows the different levels of such projects along with different parameters to compare their characteristics leveraging the capabilities in many real-life applications.

\subsection {SiriGPT}  
Siri \cite{siri} is an intelligent digital assistant that enables Apple device users to complete tasks more efficiently and with ease, often anticipating their needs even before they make requests. SiriGPT  \cite{siri1} \cite{siri2} is a voice assistant powered by a GPT model and developed entirely using shortcuts. Apple device users can utilize ChatGPT, fueled by GPT-3, by using an API key provided by OpenAI. This novel combination offers the best of both worlds, allowing users to utilize SiriGPT for voice commands and ChatGPT for generating text. SiriGPT utilizes a tokenizer exclusively developed by Apple that has been optimized for processing natural language tasks. SiriGPT's training data is not publicly available as it is exclusive to Apple. However, the language model is trained on diverse text data from various sources such as books, news articles, web pages, and other text data sources. This ensures that SiriGPT can handle different natural language tasks accurately and efficiently. It has been reported that SiriGPT is one of the largest language models available, with over a trillion parameters.

\subsection {AI Dungeon} 
Latitude, a startup based in Utah, created a groundbreaking online game called AI Dungeon \cite{aidungeon}, which showcased a novel type of collaboration between humans and machines. It is a  free-to-play, single-player, and multiplayer adventure game that caught traction within the gaming community. It combines fantasy and AI to create endless possibilities, e.g., one can take charge of a military operation to defend against aliens or become a famous detective investigating an attempted murder of the queen of the fairies. Unlike games with predetermined storylines, AI Dungeon allows you to guide the AI to generate unique characters and scenarios for your character to interact with. The game boasted about incorporating the GPT-3 text generator, but then the algorithm began producing unsettling narratives, including graphic depictions of sexual encounters involving minors \cite{wired}.

\subsection {Copy.ai}  

Copy.ai \cite{copyai} is a mighty AI startup founded by Paul Yacoubian in 2020. This project is created using GPT-3, mainly targeting business and marketing campaigns. It has the following use cases: (i) For Teams: It assists with producing customized sales copy, composing long-form articles and pages on a large scale, reusing content on various platforms, and creating product descriptions; (ii) For Emails: The AI-powered email writer takes care of the most challenging parts of marketing by creating email campaigns that are highly effective at converting leads, all with just a few clicks of a button; (iii) For Blogs: By generating content briefs and crafting one-of-a-kind SEO-focused blog articles every month, it can save a significant amount of money for the business. In addition, it's feasible to create briefs, outlines, and even initial drafts in mere minutes, which can be utilized as an excellent source of inspiration for writers to create high-quality content; (iv) Social Media: It aids in generating social media posts quickly and efficiently, allowing for a rapid expansion of the social media following. Additionally, Copy.ai includes a suite of other tools, such as a headline analyzer, a language translator, and a content rephrase. 

\subsection {Bond.AI} 

Bond.AI \cite{bondai} is a company focused on AI for financial institutions, which has a headquarters in Little Rock, Arkansas. It was established by Uday Akkarajuin in 2016 and prided itself on providing AI technology centred around human needs. This innovative project offers a product named BondBot, which is powered by Empathy Engine 3.0 and ChatGPT, to enhance the financial health of clients. It assists financial institutions and employers in promoting interconnected finance by offering various tools to improve the institution's profitability and the financial health of its clients on a single network. It uses customer data to create individual personas for every bank customer or small business, considering their behaviours, strengths, and potential needs. This approach enables the platform to develop multiple customized pathways to holistically enhance clients' financial well-being.

\subsection {Viable} 
Viable \cite{viable} is a platform powered by GPT-4 that utilizes the latest advancements in NLP and AI to offer businesses intelligent insights to aid their decision-making processes. Companies can extract actionable insights from unstructured data sources, such as social media posts, customer reviews, and survey responses, by employing Viable. GPT assists in comprehending the sentiment and context behind the data, resulting in valuable insights that can enhance a company's services, products, and customer experience. Viable's "Insight Explorer" is a distinctive feature that enables users to interact with and visualize their data via a user-friendly interface. In addition, the platform offers advanced analytics capabilities, including entity recognition, topic modelling, and sentiment analysis. The GPT-based technology of Viable is continually evolving and advancing, which allows the platform to deliver more precise and insightful data. Moreover, Viable provides customized integration and solutions to cater to the specific requirements of each business.

\subsection {AI Channels}  
AI Channels \cite{aichannels} is a platform that provides a comprehensive set of tools for developers, data scientists, and machine learning practitioners to develop, launch, and manage their AI models. The platform offers an all-in-one solution for creating personalized AI models, starting from data preparation and model training to deployment and monitoring. Users can train their models on their data or, on pre-trained models provided by AI Channels. These models can be deployed as APIs or Docker containers on various infrastructures, including different cloud platforms. It also provides a dashboard for tracking model performance and managing configurations. It covers various use cases, including computer vision, NLP, and speech recognition. The platform includes pre-built models for tasks such as image and text classification, object detection, and sentiment analysis. Additionally, users can create their models using popular frameworks. The main objective of AI Channels is to make building and launching AI models more accessible to developers and businesses without specialized AI skills. 

\subsection {Fireflies.ai} 
Fireflies AI \cite{fireflies} is a privately held company based in San Francisco, California, founded by Krish Ramineni and Sam Udotong. Fireflies AI  software is powered by GPT-4 to automate notes-taking tasks and collaborations during meetings. It is compatible with various video conferencing platforms, including Zoom, Google Meet, and Microsoft Teams, and it can transcribe meeting audio and video content in real time. Its primary function is based on speech-to-text technology, which enables it to generate a searchable transcript of the meeting, which can be used for later review and to recall essential points and action items. Additionally, the software utilizes NLP capabilities that can identify significant keywords and phrases within the conversation. Apart from note-taking, Fireflies AI includes collaboration tools such as assigning tasks and sharing notes with other team members. It can integrate with project management and task tracking tools to automatically generate tasks based on the identified action items during the meeting. Fireflies AI provides several customization options to suit particular use cases and workflows. Users can configure the software to automatically join specific meetings or capture audio only from specific speakers. It allows users to specify particular words and phrases to highlight in the transcript, making it easier to identify critical points during the later review. Thus, Fireflies AI aims to simplify the meeting process and reduce the time and energy required for note-taking and collaboration.

\subsection {Uber's Plato Research Dialogue System}   

Uber's AI Lab introduced Uber's Plato Research Dialogue System in 2020 developed by a team of researchers and engineers to enable the intelligence in riding experience. PLATO - Pre-trained Dialogue Generation Model with Discrete Latent Variable \cite{papangelis2020plato}. Uber's Plato Research Dialogue System uses GPT-2, a large-scale language model developed by OpenAI in 2019. Uber's Plato Research Dialogue System project used several datasets to train and evaluate their conversational agents such as Persona-Chat contains 160,000 conversational dialogues, Cornell Movie-Dialogs Corpus with 200,000 movie conversations, DailyDialog over 13,000 dialogues, and EmpatheticDialogues over 25,000 user dialogues. The main components in developing the GPT-powered PLATO project are
language modelling, dialogue modelling, discrete latent variable modelling and response ranking. The Plato Research Dialogue System was trained on a massive corpus of text data consisting of over 40 GB of uncompressed text while Bleu score, Perplexity and Distinct n-gram are the evaluation metrics used for training and testing the PLATO project. Uber's AI PLATO has addressed many key challenges like customer service by personalizing user feedback with conversational AI agent, user experience using the Uber platform for scheduling rides, navigating routes, and providing real-time updates, and increasing operational efficiency by reducing the need for human customer service representatives and enabling faster and more accurate communication between riders, drivers, and the Uber app. 

\subsection {Jukebox}    

Jukebox, a GPT-powered music creation, was developed in 2020 as an extension of Open AI's GPT language model \cite{dhariwal2020jukebox}. Jukebox's goal is to push the boundaries of what AI can accomplish in the world of music creation and to investigate fresh applications for AI. A variation of the GPT architecture, the "Multi-Scale Transformers for Music Modeling" (MST) model, was created specifically to handle the intricate and multi-scale nature of musical data. Additionally, Jukebox can produce lyrics that match the music's tone and style. A sizable and varied dataset of musical recordings, lyrics, and related metadata was used to train Jukebox such as 1.2 million songs sourced including Lakh MIDI Dataset, Free Music Archive, Spotify and Tidal, 600,000 pieces of sheet music were sourced from IMSLP (International Music Score Library Project), and 45,000 MIDI files from Lakh MIDI Dataset and the MIDIworld collection. Faster training times and more effective use of computational resources were made possible by the distributed computing setup with 2048 TPU( Tensor Processing Unit) cores used to train the Jukebox model. Training the model required significant computational resources demanding faster training times by the distributed computing setup with 2048 TPU  (Tensor Processing Unit) cores used to train the Jukebox model. A combination of subjective and objective metrics was used to assess and test Jukebox. In a large-scale subjective assessment, more than 1,000 participants listened and rated each one individually determining the overall score for each song produced. On the other side, objective assessments were conducted by evaluating Frechet Audio Distance (FAD) and Pitch and Rhythm Similarity. Overall, Jukebox revolutionizes with its significant advancement in the music industry through creative inspiration, music production, music education and preservation of music heritage.

\subsection {Meena}   

Google's Meena project was developed by Google Research Team in 2020 for providing personalized product recommendations \cite{adiwardana2020towards}. The primary goals of the Meena project empowered the lifestyle sector to enhance the user experience and customer service by recommending goods and services on a personalized basis. The project designed a GPT using the seq2seq transformer-based neural network architecture, in particular for open-domain conversational agents. The architecture was pre-trained over 341 GB of text captured from Reddit and other social platforms containing over 40 million words and called this massive collection as 'Meena Dataset'. Meena was tested using the automated performance metrics known as Bleu score and perplexity on a cluster of HPC nodes with a total of 2048 NVIDIA V100 GPUs. One of the biggest challenges solved Meena was building trust and generating reliable engaging human-like conservation that typically enhances user satisfaction and personalization. Meena has achieved state-of-the-art performance compared to other open-domain chatbots and revolutionized the wide range of applications in the lifestyle industry and a way beyond by providing natural and engaging responses through virtual assistants, customer service bots and personal shoppers.

\subsection {DeepScribe}

DeepScribe was a GPT-based medical project developed in 2019 by the student team at the University of California by partnering with giant US-based healthcare providers such as One Medical, Stanford Medicine, Mount Sinai and Sutter Health \cite{van2021digital}. The DeepScribe's technology aims at transcribing medical conversation allowing doctors to treat the patients rather than noting down the patient's history, enhancing the doctor-patient relationship and targeting the overall quality of patient care. Although DeepScribe used the customized variants of Open AI, the technical details of the GPT model used for customizing the model were not disclosed which was optimized for medical transcription tasks. 

\subsection {Polyglot AI}
Polyglot AI is a communication platform designed to generate text in multiple languages and process the data by performing several tasks such as advanced NLP techniques, text translation, and sentiment analysis. The potential features of Polyglot AI have been exploited in the following application areas such as language translation, chatbots, language learning tools, content creation, customer support, and data analysis across different languages and regions. Polyglot AI is built based on different variants of GPT models, and state-of-the-art language model architecture for NLP tasks, which uses the self-supervised learning approach. 

The Polyglot AI was pre-trained using a large amount of textual data on multiple languages simultaneously in an unsupervised environment using a shared architecture, Multilingual Universal Sentence Encoder (MUSE). MUSE developed by Google, is a pre-trained DL model used for cross-lingual TL, that encodes the text into common vector space for multiple languages. Thus, the Polyglot language model was created with the following pre-training techniques as Masked Language Modeling (MLM), Translation Modeling Language (TML), sequence-to-sequence modelling and cross-lingual TL. The pre-trained language model is fine-tuned and evaluated by standard benchmarks and metrics such as the BLEU score (Bilingual Evaluation Understudy), METEOR (Metric for Evaluation of Translation with Explicit ORdering) or F1-score. Remarkably, Facebook used new Polyglot AI to translate between 100 languages \cite{hao2020facebook}. Thus Polyglot AI enables absolute communication irrespective of the language barrier across different regions and cross-culturalism.

Thus, this section focused on several exciting real-life projects which are developed and used for humankind. These projects were discussed by presenting Table \ref{tab:projectsummarytable} highlighting the details of the project with model architecture, datasets used, training and testing, and evaluation metrics involved with the challenges addressed. The next section will discuss the open research issues and future directions for the potential benefits of GPT models.

\section {open research issues and future directions}
\begin{figure*}[!ht]
	\centering
	\includegraphics[width=18cm, height=20cm]{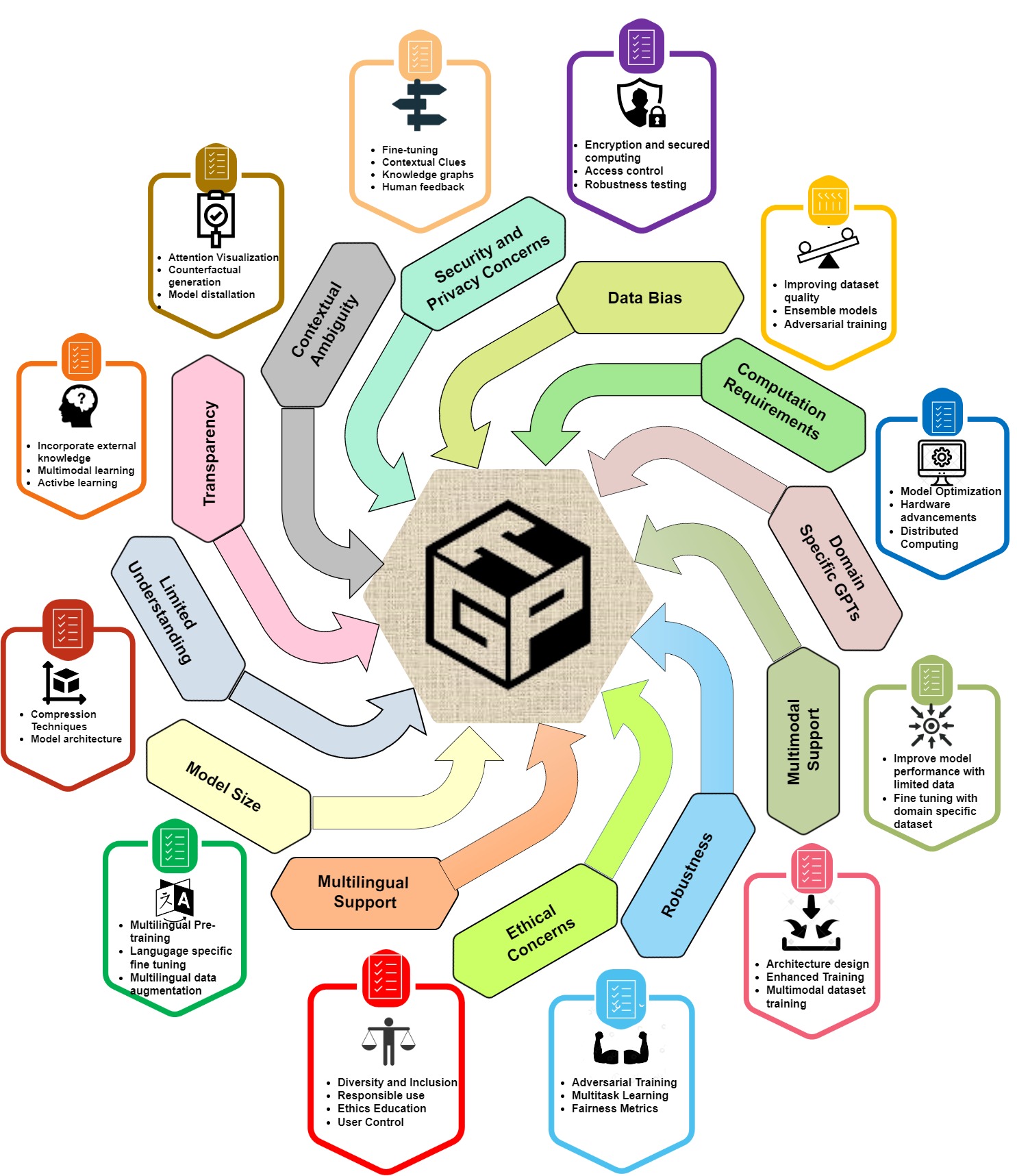}
	\caption{Challenges and Future Directions.}
	\label{fig:challenges}
\end{figure*}

This section highlights the various open research issues concerned with the implementation and adoption of sustainable GPT models. It also provides insights into future research directions for the betterment of researchers in the field of GPT development. Fig. \ref{fig:challenges} outlines the many issues that can develop while using GPT models, as well as the various future approaches that need to be considered for the effective usage of GPT models. 

\subsection{Domain Specific GPT models}
Domain-specific GPT models are mandated in almost all applications; developing these models is still challenging and an open issue within GPT. While the current GPT models have been developed to understand natural language and generate content effectively, their performance may not be equally effective when handling specific domains, such as medicine, agriculture, etc. One of the key challenges in adapting to a particular domain is the availability of domain-specific data. It is well known that the performance of GPTs is directly proportional to the quality and quantity of data used for training the model. So, obtaining such quality data for a specific domain is expensive and time-consuming, as the data are heterogeneous. Also, these data accumulations may even make these models much larger, sometimes catastrophic too, leading to forgetting the knowledge attained during the process. To overcome this issue, pre-training tasks and domain-specific model generation are integrated by data augmentation\cite{gan2023joint}.  Another challenge is fine-tuning the model to accustom to the unique characteristics and vocabulary of the domain. A few domain-specific GPT models have been developed and implemented despite these challenges. There is a growing interest in creating more domain-specific GPTs for various domains. Moreover, these models will be trained using the knowledge acquired from large language models specific to domains. Therefore, these models can be fine-tuned for specific tasks or domain-specific requirements with gradually improving performance. GPT models have the potential to be trained in any context, and researchers are exploring new approaches and methods to address these challenges. Furthermore, these models will be more efficient, enhanced interpretability, and domain generability than the existing Large language models as they are customized to specific domain concerns and can provide more concise and informative solutions. TL can be used for developing domain-specific GPT models. Domain-specific GPT models were developed to summarize products based on customer reviews on an E-commerce site, where the language model is pre-trained on the Chinese-short summarization dataset and has obtained fine-tuned results\cite{zhang2021dsgpt}. Besides these challenges, domain-specific models require higher computation costs for the resources and time spent in pre-training and relearning in downstream tasks during fine-tuning of pre-trained domain-specific models. Therefore, domain-specific model development must focus on optimizing resource consumption and fine-tuning the pre-trained model to alleviate the forgetting problem involved in existing models. 

\subsection{High Computational requirements}
 As the Transformer model utilizes varied heterogeneous datasets for training and learning from the knowledge acquired, one of the key challenges of GPT models is high computational resources for pre-training and inference. The computational requirement continuously increases as the models become more complex and larger. Depending on the size and complexity of the model and the available resources, the time required to train the model can take days, weeks, or even months. Moreover, the inference time for these models is typically slower, making it challenging to use them for real-time applications. This poses a significant obstacle to adopting GPT models for many practical applications. Despite these challenges, significant efforts are underway to overcome them. To accommodate the increasing data size and pre-training computational requirement, data enhancement-based GPT models were developed\cite{gan2023joint} by joining the downstream tasks and pertaining process by reconstructing the domain-specific text before proceeding for pre-training and utilizing the empirical knowledge rather than learning for falsy domain data. Researchers are exploring various ways to optimize and speed up the training and inference process, such as using specialized GPUs and TPUs. They are also developing more efficient algorithms and attempting to reduce the model size without sacrificing performance. In addition, ChatGPT has evolved to include plugins \cite{chatgpt_plugins_2023} that enable statistical analysis for real-time applications. By integrating these plugins with the help of third-party services, ChatGPT can now be used for analyzing real-time applications as well.
 \subsubsection{Increasing Model size and Space Constraints}
Developing and training large language models, such as a GPT model, can be a challenging task due to significant technical and computational difficulties as discussed. The size of GPT models presents a major challenge, as the computational resources required for training and inference increase with the number of parameters that need to be trained. As the model size increases, it also requires more memory to store and manipulate parameters during training and inference\cite{nori2023capabilities}. Acquiring and processing vast amounts of high-quality training data is another challenge in training large language models like GPT. For instance, GPT-4, which is the largest GPT model to date with 1 trillion parameters, demands a massive amount of computational resources, such as specialized hardware like GPUs and TPUs, spatial requirements, and high-speed network connections to transfer data between different parts of the system. Model evaluation and interpretation are also critical challenges. Since large language models like GPT are trained on a massive scale, understanding how the model makes predictions and why it generates specific outputs is difficult. Evaluating the quality and accuracy of the model's output and identifying and addressing biases or errors in its performance can also be challenging.

As these and other significant efforts continue, we can expect the challenge of computational resource requirements for GPT models to transform into a strength in the future.

\subsection{Explainability and interpretability}
Explainability and interpretability are currently major challenges for GPTs for specific applications. Explainability refers to providing a clear and understandable explanation of how the model has arrived at any output. Interpretability, on the other hand, refers to the ability to understand the internal processes of the model. GPT models are highly complex and difficult to understand and interpret due to their size and architecture. The outcomes and decisions of the model are based on previous learning and training, and the models learn from vast amounts of data to make decisions. These decisions may not be easily explainable to humans. This lack of transparency and interpretability raises concerns about the reliability and safety of the model, particularly in critical applications such as healthcare and finance. Researchers are currently conducting much research to make GPT models more explainable and interpretable \cite{bhattarai2022convtexttm} by utilizing EXplainable Artificial Intelligence (XAI) to provide explanations for the decisions arrived at, specifically to different users at stake. As well, XAI models enable interpretability by providing detailed explanations for the internal process.  As GPT can generate any type of unconstrained output for instance code generation for the given problem, it requires proper justifications and explanations for the output. So, to assure these codes by GPT are reliable, a metric model to evaluate and validate this GPT code was developed using NLP metrics and XAI for model interpretability\cite{narasimhan2021cgems}. Also, some domain-specific GPT models of GPT-3 have evolved with solutions \cite{zini2022explainability} \cite{yadav2022interpretable} to ensure that the GPT model's decisions are understandable, explainable, and trustworthy enough to be used for critical applications like healthcare and finance.

\subsection{Data Bias}
Data bias is an open issue concerned with the adoption of any advancements in AI, till GPT\cite{chan2022gpt}. This is also a prominent challenge for GPT and other machine-learning models. It refers to patterns or relationships in the data that do not accurately reflect the true distribution of the target population or domain. GPT models are trained on vast amounts of text data which may contain bias in language use or cultural assumptions. Still, the source of data remains undeclared, considering GPTs are trained using internet data which may have faulty, fake, and error data, GPTs may generate biased texts or information imitating the training data\cite{zhang2021commentary}. Such biases can be amplified in the model's output, resulting in false or unfair results. Data bias can arise from various sources, such as selection bias, labelling bias, concept drift, confounding variables, and changes in input data distribution over time. For example, suppose a dataset used to train a GPT model is dominated by a particular demographic group. In that case, the resulting model may be biased in its predictions towards that group, leading to inaccurate or unfair predictions when applied to new data. This bias can have serious consequences, especially in healthcare, finance, and law enforcement, where biased results can significantly impact human lives. To mitigate these issues, researchers have developed strategies such as diversifying the training data, debiasing the training data, modifying the model architecture, and using post-processing methods to normalize the data and create more fair and inclusive GPT models. The authors in \cite{kirk2021bias} have made an in-depth analysis of the most downloaded text generation model GPT2. By examining the intersections of gender with religion, sexuality, ethnicity, political affiliation, and continental name origin, the authors evaluated prejudices associated with occupational associations among various protected categories. These biases may have inaccuracies in climatic data prediction or global warming\cite{biswas2023potential}. Therefore, data bias must be of greater concern in GPT model development as the data quality of the internet is limited to avoiding producing disturbing content.

\subsection{Multimodal support}

The challenge of developing multimodal learning ability in the GPT model remains unsolved. Multimodal support refers to the GPT model's ability to process and generate text along with other modalities, such as audio, images, and videos. GPT models have shown impressive results in generating high-quality text and NLP tasks, but it was primarily designed for text-based tasks and cannot handle other modalities. However, due to its success in text processing, users expect its integration with other modalities, such as speech recognition, video summarization, and image or video captioning\cite{seo2022end}. Several research initiatives have been proposed to integrate multimodal support to address this issue. One approach is to feed the visual and audio information with the corresponding text to the model as input. The other is to handle this input modality process as a separate model and use the output as input to GPT. Multimodal video captioning is done using GPT in the unlabelled videos\cite{seo2022end}. Multimodel learning has been applied for information retrieval\cite{ji2023deep} and image generation for illustrating the news\cite{liu2022opal} to assist the GPTs. However, the primary challenge in both approaches is effective integration, requiring architectural changes and techniques to handle various modalities. Recently, OpenAI's GPT4 has launched with multimodal support, enabling it to read images, analyze the input, and generate text as output. It cannot create images as output, though. Nevertheless, the field of multimodal processing is still an active area of research, and much work must be done to effectively and efficiently process and understand multimodal data. 

\subsection{Robustness}

The robustness is a major requirement to be imposed by any type of GPT model, and it is a global problem for all learning-based prediction technologies. Robustness refers to the ability of the model to maintain high performance and accuracy even in the face of unexpected or adversarial inputs. Although GPT models have shown impressive performance in a wide range of NLP applications and have set a benchmark for high-quality text generation, they are still vulnerable to certain types of errors and attacks. In particular, handling adversarial inputs is a challenging task in GPT models. GPT models are particularly susceptible to adversarial attacks\cite{guo2021gradient}. Adversarial inputs are specifically designed to make a learning model collapse and misbehave. GPT models can be highly prone to these attacks because they are trained on a large volume of text. As a result, they may be influenced by subtle patterns or biases in the training data. If such biases or patterns exist in the data, the GPT model may amplify or perpetuate existing biases, leading to unfair outcomes. A few techniques may be used, such as adversarial training \cite{DBLP:journals/corr/abs-1904-12843}\cite{DBLP:journals/corr/abs-2102-01356}, defensive distillation \cite{article1}, and regularization techniques \cite{inbook} such as dropout, weight decay, and batch normalization, to mitigate and handle adversarial inputs. Therefore, GPT development must focus on developing models with more robustness, enabling them to be tolerant of various vulnerabilities, and thus to be used reliably and susceptible in a wide range of applications.

\subsection{Multilingual support}
While GPT models have demonstrated remarkable proficiency in NLP tasks for individual languages, achieving multilingual support remains a significant challenge. The primary difficulty in developing multilingual GPT models lies in the significant differences in syntax, grammar, and vocabulary across various languages. As the number of internet users day by day increasing irrespective of literacy rate, multilingual support will target all types of end users. To create models that can effectively process multiple languages, researchers need to train GPT models on extensive, diverse datasets that span a broad range of languages and language families. Additionally, designing language-specific pre-processing techniques to prepare input data for the model is another obstacle to overcome. Various languages possess distinct writing systems, word orders, and linguistic features, necessitating specialized pre-processing techniques to ensure that the model can process the input data effectively. Despite the challenges, researchers continue to explore new methods to improve the multilingual capabilities of GPT models. Some techniques involve training separate models for each language or developing language-specific fine-tuning techniques. Others include developing cross-lingual TL techniques that allow the model to transfer knowledge and skills learned in one language to another. 

\subsection{Limited understanding}

GPT models have a limited understanding of context and meaning, despite their ability to generate coherent text. This problem arises due to issues such as a lack of semantic understanding, bias, stereotyping, and handling nuances and figurative language. As a result, the outputs generated by the model may contain errors or inaccuracies, even if they are grammatically correct. Researchers are exploring various techniques to enhance the model's contextual understanding. Understanding GPTs will be more reactive and may attract more users for accurate results\cite{liu2021gpt}. These methods include incorporating external knowledge sources like knowledge graphs and ontologies into the training process, developing common sense reasoning capabilities, and improving the model's ability to handle nuances and idiomatic expressions. By enhancing the contextual understanding of GPT models, their outputs will be more accurate, relatable, sequential, less biased, and more useful for a variety of applications.

\subsection{Ethical Concerns}
The ethical concerns in GPT models are an active area of discussion and debate due to the potential negative impacts that the use of GPTs could have on society. Although GPT models have demonstrated remarkable abilities in generating coherent and realistic text, there are concerns about the perpetuation of biases and stereotypes, the possibility of malicious use, and the effects on employment and economic inequality. Some of the ethical characteristics to be possessed by GPT include functional Morality, operational morality, abiding by the right for explanation law, improved transparency with human involvement, unbiased data, and adhering to government regulations on data usage\cite{chan2022gpt}.  The responsibility of developers and companies to address these ethical concerns and ensure the ethical use of a GPT model is also a topic of debate. The ethical implications of GPT models are being actively researched and discussed in the fields of AI, computer science, and philosophy.

\subsection{Security and privacy concerns}
GPT models raise concerns about security and privacy, particularly as they become more widespread. One of the main concerns is that GPT could be used for harmful purposes, such as creating fake news or deep fakes, as it can generate text that looks real and convincing, making it difficult to distinguish between genuine and fake content. Another concern is the potential for privacy violations when using a GPT model. Large language models like GPT require a significant amount of training data, which could contain sensitive or personal information. This raises concerns about privacy and data protection as per European Union's General Data Protection Act\cite{jia202110}, particularly if the training data is not properly anonymized or if the models are used to generate text based on user data without their explicit consent. Some of the problems concerned with confidentiality related to the pre-training dataset are Data tracing, Membership Inference Attacks, reconstruction attacks, and property inference attacks and the vulnerabilities concerned with a model encoder are hyperparameter stealing attacks and encoder parameter stealing attacks. Poisoning, Backdoor, and evasion attacks are the vulnerability related to the integrity of self-supervised learning. Resource depletion attack is one major issue with data availability, which may lead to tremendous effects incorrect results, and may cause greater deviations too\cite{jia202110}.   Additionally, the GPT model's ability to generate text based on user input could inadvertently disclose sensitive information, such as personal or financial details, or trade secrets. This could happen if a GPT model is used in an insecure environment or if it is targeted by malicious actors seeking to obtain sensitive information. Researchers and developers should focus on assuring authenticity in using users' data in case of interactive information generation based on privacy data shared. These include using differential privacy to protect training data privacy \cite{abadi2016deep}, implementing secure hardware or software protocols to protect models from cyberattacks, and developing techniques to detect and prevent the malicious use of GPT models. It's crucial to adopt and follow these measures to ensure the ethical and safe use of GPT models before using them in various applications.

Therefore, GPT model development must focus on developing more robust, reliable, safest, multi-lingual, multimodal support-enabled solutions for delivering domain-specific or human-specific solutions with optimal resource utilization.
\begin{table*}[!ht]
\centering
\caption{Various lessons learned and future research directions.}
\label{tab:lessons}
\resizebox{\textwidth}{!}{%
\begin{tabular}{|l|p{3.5cm}|p{6cm}|p{6cm}|}
\hline
Sl.No &
  Lessons Learned &
  Open Issues &
  Future Directions \\ \hline
  1. &
  $\circ$ Huge volume of data usage is critical
 &
    
  $\circ$ Data privacy - may unknowingly reveal sensitive information
  
  $\circ$ Varied data quality - Inconsistency in quality of data used for training 
  
  $\circ$ Scalability - Models should be able to handle an increase in data set size and complexity &
  $\circ$ Optimized architecture and algorithms 
  
  $\circ$ Cloud-based computing
  
  $\circ$ Hardware advancements
  
   \\ \hline
 2.&
  $\circ$ Importance of Proper Pre-processing of data 
  
  &
  
  $\circ$ Data bias - Overrepresentation of certain groups or perspectives 
  
  $\circ$ Poor model performance 
  
  $\circ$ Reduced efficiency of the model 
  &
  $\circ$ Continuous monitoring
  
  $\circ$ Testing model for potential biases
  
  $\circ$ Diversifying the training data
  
  \\ \hline

  3. &
  $\circ$ Importance of explainability and interpretability
 &
    
 $\circ$ Complexity of models
  
 $\circ$ Inability to explain predictions
 
 $\circ$ Need of User-tailored Explanations generation
 
 $\circ$ Developing Interpretable models
 
  $\circ$ Lack of transparency in the data source
  
  &

  $\circ$ AI governance models can be used
  
  $\circ$ Model Summaries can be provided
  
  $\circ$ Techniques like LIME(Local Model-Agnostic Explanations can be used
  
  $\circ$ Uncertaining estimates can be obtained from a model
  
   \\ \hline

    4. &
  $\circ$ Ethical concerns
 &
    
  $\circ$ Data privacy and data protection
  
  $\circ$ Misuse of data

  $\circ$ Accountability  and transparency concerns

  $\circ$ Societal implications - displacing jobs and exacerbating equalities

  &

  $\circ$ Counterfactual analysis can be used
  
  $\circ$ Federated learning can be used
  
  $\circ$ Ethical guidelines, Legal frameworks and regulations can be developed to avoid harmful use
    
   \\ \hline

   5.  &
  $\circ$ Lack of contextual understanding in AI systems
 &
    
  $\circ$ Possibility for ambiguous, contradictory, incorrect results leads to misunderstandings
  
  $\circ$ Inconsistency in responses or outputs

  $\circ$ Lack of ability in distinguishing true and false information

  &

  $\circ$ Incorporation of knowledge graphs and semantic embeddings into the training process 
  
  $\circ$ Usage of attention mechanisms to focus on relevant parts of the input
  
  $\circ$ Imparting reasoning and inference capabilities

  $\circ$ Task or domain-based fine-tuning
    
   \\ \hline

   6. &
  $\circ$ Pre-trained models may not perform well for Domain-specific task
 &
    
  $\circ$ Possibility for ambiguous, contradictory, incorrect results leads to misunderstandings
  
  $\circ$ Inconsistency in responses or outputs

  $\circ$ Lack of ability in distinguishing true and false information
  
  &

  $\circ$ Incorporation of knowledge graphs and semantic embeddings into the training process 
  
  $\circ$ Usage of attention mechanisms to focus on relevant parts of the input
  
  $\circ$ Imparting reasoning and inference capabilities

  $\circ$ Task or domain-based fine-tuning
    
   \\ \hline
\end{tabular}%
}
\end{table*}

\section{Conclusion}
 The impact of GPT and other large language models is far-reaching and profound. As these technologies continue to evolve and improve, they have the potential to transform the way we interact with technology and each other. From personalized recommendations and customer service to language translation and text generation, the possibilities are endless. However, as with any technology, there are potential ethical and societal concerns that must be addressed. As we continue to rely more heavily on these language models, we must ensure that we are using these tools responsibly and with consideration for their impact on society as a whole. These include challenges related to biases in the data used to train the models, safeguarding privacy and security, understanding the implications of human creativity, and the potential impact on employment and job displacement. We need to continue to evaluate and reflect on the impact of GPT and other language models, to ensure that they are being used in a way that benefits society as a whole. By doing so, we can help to ensure that these technologies are used to their fullest potential while minimizing any negative impact that they may have.


\balance
\bibliographystyle{IEEEtran}
\bibliography{citation}

\end{document}